\documentclass[11pt]{article}
\pdfoutput=1
\usepackage{acl}
\usepackage{times}
\usepackage{latexsym}

\usepackage{algorithm}
\usepackage{algpseudocode}
\usepackage{amsmath}
\usepackage{amssymb}
\usepackage{booktabs}
\usepackage{bm}
\usepackage{caption}
\usepackage{cleveref}
\usepackage{enumitem}
\usepackage{graphicx}
\usepackage{listings}
\usepackage{longtable}
\usepackage{mathtools}
\usepackage{placeins}
\usepackage{subcaption}
\usepackage{tcolorbox}
\usepackage{verbatim}
\usepackage{wrapfig}
\usepackage{multirow}
\usepackage{pythonhighlight}

\newcommand{\bfc}{\mathbf{c}}

\newcommand{\bfq}{\mathbf{q}}

\newcommand{\bfp}{\mathbf{p}}

\newcommand{\bfz}{\mathbf{z}}

\newtheorem{definition}{Definition}
\newtheorem{proposition}{Proposition}
\newtheorem{proof}{Proof}
\usepackage[T1]{fontenc}

\usepackage[utf8]{inputenc}

\usepackage{microtype}

\usepackage{inconsolata}

\usepackage{todonotes}

\title{Calibrating Verbalized Probabilities for Large Language Models}
\author{
  \textbf{Cheng Wang, Gyuri Szarvas, Georges Balazs} \\
  \textbf{Pavel Danchenko, Patrick Ernst} \\
  \texttt{\{cwngam, szarvasg, jabalazs, danchenk, peernst\}@amazon.com} \\}

\begin{document}
\maketitle

\begin{abstract}
Calibrating verbalized probabilities presents a novel approach for reliably assessing and leveraging outputs from black-box Large Language Models (LLMs). Recent methods have demonstrated improved calibration by applying techniques like Platt scaling or temperature scaling to the confidence scores generated by LLMs. In this paper, we explore the calibration of verbalized probability distributions for discriminative tasks. First, we investigate the capability of LLMs to generate probability distributions over categorical labels. We theoretically and empirically identify the issue of \textit{re-softmax} arising from the scaling of verbalized probabilities, and propose using the \textit{invert softmax trick} to approximate the "logit" by inverting verbalized probabilities. Through extensive evaluation on three public datasets, we demonstrate: (1) the robust capability of LLMs in generating class distributions, and (2) the effectiveness of the \textit{invert softmax trick} in estimating logits, which, in turn, facilitates post-calibration adjustments.
\end{abstract}

\section{Introduction}
Large Language models (LLMs), such as GPT-4~\citep{openai2023gpt}, Claude\footnote{https://www.anthropic.com/claude}, Mistral~\citep{jiang2023mistral} have demonstrated to be versatile tools with strong capabilities across many natural language processing (NLP) tasks. 
Their performance is grounded on very large numbers of parameters and the refinement to align more closely with human preferences using reinforcement learning from human feedback (RLHF;~\citealp{10.5555/3495724.3495977}), direct preference optimization (DPO;~\citealp{rafailov2024direct}), etc. 
This provides LLMs with the flexibility to adapt to new tasks through natural language instructions (prompts) and contextual data (in-context learning) without requiring further training or parameter updates.

 \begin{figure}
	\centering
   \includegraphics[width=0.5\textwidth]{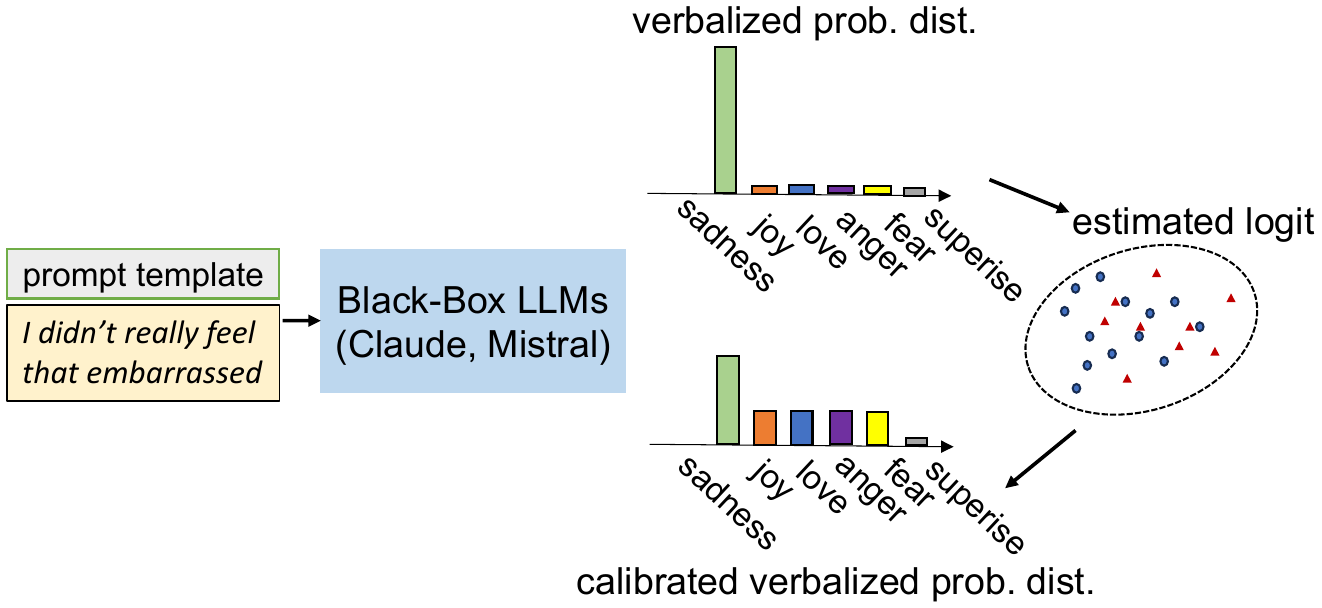}
   \caption{Our proposed method for calibrating verbalized probability based on invert softmax trick. The LLMs generated probabilities are inverted to estimate the logits, and then post-hoc calibration --- temperature scaling is applied thereafter to obtain calibrated probabilities.}
   \label{fig:workflow}
  \end{figure}

While these LLMs achieved impressive performance on a wide range of tasks, recent studies suggest that the RLHF fine-tuned LLMs are poorly calibrated~\cite{tian-etal-2023-just, zheng2023large, zhao2021calibrate, xiong2023can}. 
This prevents a reliable assessment of the confidences of LLM outputs, which is a key requirement in risk-sensitive domains such as medicine~\citep{10399826} and finance~\citep{bahnsen2014improving}, where one wants to prefer high-confidence decisions and avoid low-confidence actions. Unfortunately, calibrating LLMs is challenging, particularly for black-box LLMs, which only provides closed-source APIs and thus users have no access to model output such as token-level probabilities.

Recently, \citet{tian-etal-2023-just} proposed prompting large language models (LLMs) to directly generate prediction confidence scores, a concept they refer to as \textit{verbalized confidence}~\cite{lin2022teaching}. In this approach, the language model is instructed to respond with both the predicted label and a numeric confidence score within the range of [0, 1], or with a natural language expression indicating confidence, such as "highly likely." This procedure is known as \textit{confidence elicitation}.

Their empirical results show that these verbalized probabilities offer better calibration than conditional probabilities—specifically, those obtained directly from the token-level probabilities during LLM inference. Additionally, they found that fitting a single scaling parameter and using such scaled probabilities further improved model calibration, as measured by standard calibration metrics. Another concurrent study by \citet{xiong2023can} extended the work of \citet{tian-etal-2023-just}. This study combined verbalized confidence with sampling strategies to generate multiple responses and an aggregation approach to ensure response consistency.

While these approaches have shown promising performance calibrating verbalized confidence that associates to a response, in this work, we extend the scope to verbalized probability distribution for dismincriative tasks. We try to explore and answer the following open questions:
\begin{enumerate}[label=(\roman*)]\itemsep-1.5mm
  \item Can we generalize confidence elicitation to probability elicitation for discriminative tasks that require a full probability distribution over all categorical classes? That is, instead of generating prediction and associated confidence score, can LLMs generate probability distribution?
  \item For classification tasks, the LLM generated probabilities are analogous to conventional ``softmax probabilities''. however, in post-hoc calibration method, for example, \textbf{temperature scaling (TS)\footnote{Same to Platt scaling and vector scaling~\citep{guo2017calibration}}: $\bfp = \textsc{softmax}(\bfz/\tau)$ is normally used to scale unnormalised logits $\bfz$ provided by the neural network, and not softmax probabilities directly}. What is the consequence of scaling the generated probabilities directly for classification?
\end{enumerate}

\begin{table*}[!htb]
\centering
\small
\caption{Examples of LLMs generated probability distribution over classes.}
\vspace{-4mm}
\begin{tabular}{p{1cm}p{6cm}p{6.5cm}}\\ \toprule
\textbf{Dataset}&\textbf{Text} & \textbf{LLM Generated Probability Distribution}   \\ \midrule
IMDB &``...Everyone should take a look at this movie" & \{'positive':0.99, 'negative':0.01\} \\
Emotion &``im feeling quite sad and sorry for myself but ill snap out of it soon" & \{'sadness':0.8, 'joy':0.05, 'love':0.05, 'anger':0.05, 'fear':0.05, 'surprise':0.0\} \\
Massive &``lower all volume on speakers please" & \{`audio\_volume\_down':0.99,audio\_volume\_mute':0.01, 'audio\_volume\_other':0.0, ...\} \\ \bottomrule
\end{tabular}
\label{tab:example_text}
\vspace{-4mm}
\end{table*}

To explore the first question (i), we crafted prompt templates to request LLMs to provide probability distributions (examples are shown in Table~\ref{tab:example_text}). 
For the second question (ii), we first identify the issue of performing TS over verbalised probability (versus in the logits space), and propose an approach which estimates the logit from verbalized probability, we outline the procedure in Figure~\ref{fig:workflow}.
To summarize, our work makes the following contributions: 
\begin{enumerate}
\itemsep-1.5mm
\item We explore the capability LLMs have to generate probability distributions for discriminative tasks. To our knowledge, this is the first work to investigate this problem (Sec.~\ref{sec:prompt}). 
\item We theoretically and empirically identify the issue that is caused by re-softmaxing the generated verbalised probability when applying popular post-hoc calibration--temperature scaling(Sec.~\ref{sec:ts_prob}).  
\item To address this (ii), we use the \textit{invert softmax trick}, which inverts probabilities to estimate and approximate logits. Then performing temperature scaling to avoid the re-softmax issue(Sec.~\ref{sec:inv_softmax_trick}).
\item We conduct extensive experiments and an ablation study on three datasets to understand the TS-based calibration behaviors for discriminative tasks(Sec.~\ref{sec:experiments}).
\end{enumerate}

\section{Methods}
We describe our proposed approach to calibrate LLMs generated verbalised probability for discriminative tasks. First, we introduce a prompt for generating probability distributions over categorical labels; Then, we further identify the issue of applying post-hoc calibration method -- temperature scaling (TS) -- to verbalised probability, both theoretically and empirically. Last, we demonstrate the approach of estimating logits from the verbalised probability. 

\subsection{Prompting to Generate Probability Distributions}
\label{sec:prompt}
We used the following prompt template to explicitly ask LLMs to output probability distributions over binary labels. 
We adapted the basic prompt structure from ~\cite{tian-etal-2023-just} and tried to guide the LLM to reach our goal step by step~\cite{xiong2023can}. This empirically provides better performance in terms of both accuracy and more structured responses. We started with an easier step by asking the LLM to give a label, then asked it to consider the confidence of this label, and finally to give the full probability distribution over labels. This creates an order of incremental difficulty, where each step builds on the previous one. Lastly, we added constraints to ask the LLM to output only the probability (the part we are interested in). See \cref{sec:appendix:prompts-1,sec:appendix:prompts-2} for the exhaustive list of prompts we used. Table~\ref{tab:example_text} presents the output probability distribution. 
\begin{tcolorbox}
\begin{lstlisting}[basicstyle=\ttfamily\footnotesize]
Give a binary sentiment label
(positive or negative) to the 
following sentence: \$text.

Assign a confidence score (between 0
and 1.0) to this prediction.
Give ONLY the probability
distribution over the sentiment
labels.

Give ONLY the probability, no other
words or explanation.

Provide ONLY the probability in a
format of Python dict.
\end{lstlisting}
\end{tcolorbox}

\begin{figure*}[!htb]
\centering
   \subfloat[\scriptsize{LLM generated probability $\bfp$}]{{\includegraphics[width=0.265\textwidth]{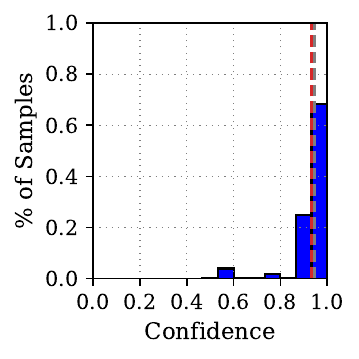} }} 
 \subfloat[\scriptsize{re-softmaxing probability from $\bfp$ to $\bfq$}]{{\includegraphics[width=0.265\textwidth]{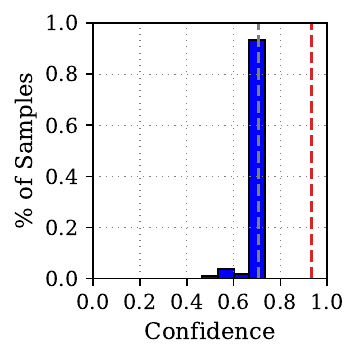} }} 
    \subfloat[\scriptsize{TS calibration based on $\bfq$ ($\tau_q= 0.31$)}] {{\includegraphics[width=0.265\textwidth]{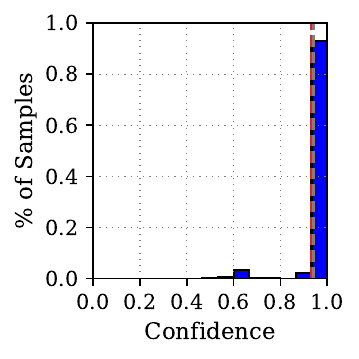} }} 
\caption{Probability histogram for (a) LLM generated verbalised probability, (b) re-softmaxed probability, (c) TS on the re-softmaxed probability. The \textcolor{red}{\textbf{red dashed}} and \textcolor{gray}{\textbf{gray dashed}} represent \textit{Accuracy} and \textit{Average Confidence} respectively. Their distance represents miscalibration, i.e. the gap to perfect calibration. The (b) and (c) steps are internal stages of TS.}
\label{fig:motivation_example}
\end{figure*}

\subsection{Temperature Scaling with Verbalized Probability}
\label{sec:ts_prob}
Temperature Scaling (TS; \citealp{guo2017calibration}) is one of the most representative post-hoc calibration methods. It uses a temperature parameter $\tau$ to calibrate the softmax probability:
\begin{equation}
p_i= \frac{\exp (z_i/\tau)}{\sum_{j=1}^{k} \exp (z_j/\tau)}, ~~i \in [1\dots k]
\label{eq:diffrentiable_approxmiation}
\end{equation}
where $\tau>0$ is used as a scaling factor that is applied to unnormalised activation value (logits) $z_i, i \in [1\dots k]$. TS controls the model's confidence by adjusting the sharpness of distribution so that the model prediction is not too certain (overconfident) or too uncertain (underconfident).

In the context of black-box LLMs, access to logits and raw probabilities is unavailable. While LLMs can generate verbalized probabilities, requesting them to produce logits is unrealistic due to the unbounded nature of logits. For discriminative task, we argue that directly performing TS on LLM generated verbalized probabilities lead to re-softmaxing.

\begin{definition}
  Re-softmaxing is a sequential operation of applying \textsc{softmax} function twice on the same data $\bfz=\{z_1,...,z_i,...z_K\}$, i.e., $\bfp=\textsc{Softmax}(\bfz)$, $\bfq = \textsc{Softmax}(\bfp)$.
\end{definition} 
Let's first understand the effects of re-applying the softmax function (i.e. \textit{re-softmaxing}) towards temperature scaling (TS)-based calibration

\begin{proposition}
\label{proposition_1}
Given a categorical probability distribution $\bfp = \{p_1,...,p_i,...,p_K\}$ over $K$ ($K>1$) classes, and a re-softmaxed probability distribution $\bfq = \textsc{softmax}(\bfp) = \{q_1,...,q_i,...,q_K\}$, for $i\in [1, K]$, the $q_i$ is bounded: $0<\frac{1}{K-1+e} \leq q_i \leq \frac{e}{K-1+e}< 1$
\end{proposition}

\begin{proposition}
\label{proposition_2}
In the context of calibrating verbalized probability using temperature scaling (TS), let \(\tau_p\) and \(\tau_q\) represent the optimal temperature values for calibrating \(\bfp\) and \(\bfq\) respectively. It follows that \(\tau_q < \tau_p\).
\end{proposition}

The proofs can be found in the Appendix~\ref{proof:prop_1}. Essentially, the proposition~\ref{proposition_1} indicates that re-softmaxing directly on verbalised probability produces a more uniformed distribution as compared to the original predictive probability distribution; The proposition~\ref{proposition_2} demonstrates that calibrating re-softmaxed probability leads to smaller temperature than expected temperature for calibrating the original probability. This causes unreliable calibration procedure because the distorted probability from re-softmaxing.

Empirically, we examine the effect of applying Temperature Scaling (TS) directly to verbalized probabilities using the IMDB dataset~\cite{maas-EtAl:2011:ACL-HLT2011}, with the validation set described in Table~\ref{tab:post_processed}. Additional empirical results are reported in Table~\ref{tab:agg_result_t0_with_resoftmax} in Appendix. Figure~\ref{fig:motivation_example} (a) shows that the generated verbalized probability tends to be overconfident (average confidence > accuracy). Calibrating this probability with TS typically requires an optimal temperature $\tau > 1$ and $\tau \to \infty$ to flatten the distribution, thereby making it less confident. However, Figure~\ref{fig:motivation_example} (c) indicates that the optimal temperature $\tau_q < 1$, which is supposed to calibrate an underconfident probability distribution. This occurs because the re-softmaxing (Figure~\ref{fig:motivation_example} (b)) results in $\bfp$ becoming a more uniform (underconfident) distribution, necessitating $\tau < 1$ and $\tau \to 0$ for proper calibration~\cite{guo2017calibration,wang2019state}. 

Based on theoretical considerations and empirical findings, \textbf{it is clear that avoiding the re-softmaxing issue is crucial to ensuring a reliable calibration procedure when applying temperature scaling to LLM-generated verbalized probabilities on discriminative tasks}. To that end, we propose to \textit{invert Softmax Trick} which reverts verbalized probabilities into estimated logits, which can then unitized in TS.

\subsection{Invert Softmax Trick}
\label{sec:inv_softmax_trick}
The standard softmax takes activation value (logits) $\bfz = \{z_1, z_2,...,z_K\}$ as input and returns a probability distribution $\textbf{p}=\{p_1, p_2, ..., p_K\}$ over $K$ classes:
\begin{align}
   \textsc{softmax}(z_i) = p_i = \frac{e^{z_i}}{\sum_{j=1}^{K} e^{z_j}}
\label{eq:softmax}
\end{align}
For a given softmax probability distribution $\textbf{p}$, reverting eq.(\ref{eq:softmax}) to obtain exact raw logits is not possible because the softmax is many-to-one (non-bijective) function. Thus we propose to obtain a logit proxy by inverting the softmax probability~\citep{smith2017offline,artetxe2018generalizing,balanya2024adaptive}. Here, we follow the implementation in ~\cite{pavpackage}:
\begin{align}
\textsc{inv\_softmax}(p_i) &= z_i = \log p_i + c\\ s.t. ~~
c &= -\frac{1}{K}\sum_i \log p_i
\label{eq:inv_softmax}
\end{align}
where $c$ is a constant scalar. In practice, we found that the selection of $c$ doesn't affect any calibration metrics (that is, we set $c=0$ or $c=1$ and obtained the same results). Then calibration can be performed by scaling the logit proxies:
\begin{align}
\hat{\bfp} = \textsc{softmax}(\textsc{inv\_softmax}(\bfp)/\tau)
\label{eq:inv_softmax_cal}
\end{align}
where $\hat{\bfp}$ is a calibrated probability from $\bfp$. $\tau$ is tuned on the validation set by minimizing the Negative-Log Likelihood(NLL) or Expected Calibration Error(ECE).

\noindent \textbf{Discussion}. The LLM generated probability distribution over classes $\{p_1, p_2, ..., p_K\}$ is supposed to be subject to $\sum_i^K p_i=1$. In practice, depending on individual capability, as shown in Section~\ref{sec:gen_prob_dist}, some LLMs fail to produce probability distributions that sum to 1, which leads to unreliable probability calibrations. Equation (\ref{eq:inv_softmax_cal}) brings the additional advantage of re-normalizing the inverted probabilities (logits) before performing calibration.

\section{Experiments}\label{sec:experiments}
Our experiments focus on multi-class classification task in a ~\textbf{zero-shot learning} setup. This is a general setup where we prompt LLMs without any fine-tuning steps involved. More specifically, we design experiments to accomplish the following objectives:
\begin{itemize}\itemsep-1mm
\item Examine the capability of LLMs in generating probability distributions on discriminative tasks in a zero-shot inference setup.
\item Assess the effectiveness of applying temperature scaling with the invert softmax trick when calibrating LLMs with temperature scaling on verbalised probabilities.
\item Understand the properties of calibrated verbalised probability in controlling Precision-Recall threshold.
\end{itemize}

\noindent \textbf{Datasets}.
In our experiments we used three text classification datasets that are available for research purposes and publicly hosted on Huggingface:

\begin{itemize}\itemsep-1mm
\item \textbf{IMDB}\footnote{\url{https://huggingface.co/datasets/stanfordnlp/imdb}}~\citep{maas-EtAl:2011:ACL-HLT2011} is a binary classification dataset for sentiment analysis. It has 25k train and test instances respectively. We sampled 5k random instances from the training set to create the validation set and used the full test set for evaluation.
\item \textbf{Emotion}\footnote{\url{https://huggingface.co/datasets/dair-ai/emotion}}~\citep{saravia-etal-2018-carer} is a 6-way classification task and has 16k/2k/2k train/validation/test splits respectively.
\item \textbf{Amazon massive}\footnote{\url{https://huggingface.co/datasets/AmazonScience/massive}}~\citep{fitzgerald2022massive,bastianelli-etal-2020-slurp} consists of parallel sentences from 52 languages. We use only the \textit{en-US} subset, which contains 11.5k/2.03k/2.97k train/validation/test splits. We use this subset for intent classification tasks, which require a model to classify a given input into one of 60 intents.
\end{itemize}

\noindent \textbf{Evaluation Metrics}
We measure predictive performance by using \textit{Accuracy}, and calibration performance by using negative-log likelihood (NLL), Expected Calibration Error (\textit{ECE}), Maximum Calibration Error (MCE) (\textit{MCE}) and Reliability Diagram~\cite{NaeiniETAL:15, guo2017calibration}.  

\textit{Expected Calibration Error} (ECE) quantifies the discrepancy between predicted probabilities and observed frequencies by dividing the prediction space into several bins $\{b_1, b_2,...,b_M\}$. Within each bin, it calculates the average absolute difference between the predicted probabilities and the accuracy:
$\textsc{ECE} =\frac{1}{N} \sum_{m=1}^{M} \left | b_m \right | \left |\textrm{Acc}(b_m) - \textrm{Conf}(b_m) \right |$ where $N$ is the total number of samples. $\left | b_m \right |$ is the number of samples in bin $b_m$, and $\textrm{Acc}(b_m) =\frac{1}{\left|b_m\right|}\sum_{i \in B_m}\mathbf{1}(\hat{y}_i=y_i)$, $\textrm{Conf}(b_m) =\frac{1}{\left|b_m\right|}\sum_{i \in b_m}p_i$.

\textit{Maximum Calibration Error} (MCE) focuses on identifying the largest discrepancy between predicted probabilities and empirical accuracy within any individual bin,
$
\textsc{MCE} = \max_{m\in \{1, \dots, M\}} | \textrm{Acc}(b_m) - \textrm{Conf}(b_m)|
$
and is particularly important in high-risk applications where reliable confidence measures are absolutely necessary. 

\noindent \textbf{Implementation Details}.
We used three black-box LLMs: Claude-v2\footnote{\texttt{anthropic.claude-v2}}, Claude-v3\footnote{\texttt{anthropic.claude-3-sonnet-20240229-v1:0}} and Mixtral 8$\times$7B\footnote{\texttt{mistral.mixtral-8x7b-instruct-v0:1}} through AWS Bedrock\footnote{\url{https://aws.amazon.com/bedrock/}}, and performed prompting to obtain probability distributions over categorical labels. The prompting templates are in \cref{sec:appendix:prompts-1,sec:appendix:prompts-2}. In practice, LLMs can return unexpected responses for some inputs. We observed two main reasons for such cases: (1) some LLMs may refuse to express sentiment or emotion to some input sentences from IMDB and Emotion datasets due to their internal ethical guardrails; (2) they may not be following prompting instructions precisely as mentioned in Section~\ref{sec:prompt}, for example, they may still generate extra text content beyond the probability distribution thus invalidating output parsing. In these cases, we re-ran the same or slightly modified prompt\footnote{by changing ``\textit{no} other words" to ``\textit{remove} other words"} multiple times until we saw no further improvements. During validation and evaluation we ignored the instances that still showed these issues.

Table~\ref{tab:post_processed} (top) outlines the statistics of post-processed LLM predictions. Note that the small difference in the numbers of post-processed validation/test samples makes direct and exact comparison between LLMs less fair. With this group of datasets, we compare the calibration performance against the uncalibrated baseline. The results are summarised in Table~\ref{tab:result}. To make a more direct and fair comparison across LLMs we aggregated the val/test subsets by finding the maximum intersection across LLMs (see Table~\ref{tab:post_processed}; bottom), and report the performance in Table~\ref{tab:agg_result}.

\begin{table}[!htb]
\centering
\small
\caption{The post-processed LLM predictions. The aggregated datasets are the predictions of the max intersection (based on input text) from three LLMs.}
\label{tab:post_processed}
\setlength{\tabcolsep}{8pt}
\begin{tabular}{llcc} 
\toprule 
Datasets & LLMs & Validation & Test \\ 
\midrule 
\multirow{3}{*}{IMDB} & Claude-v2 & 4998 & 24998 \\ 
 & Claude-v3 & 5000 & 24988 \\
 & Mixtral & 5000 & 24999 \\\cmidrule{2-4}
\multirow{3}{*}{Emotion} & Claude-v2 & 1988 & 1993 \\
 & Claude-v3 & 2000 & 2000 \\
 & Mixtral & 1988 & 1984 \\\cmidrule{2-4}
\multirow{3}{*}{Massive} & Claude-v2 & 2030 & 2967 \\
 & Claude-v3 & 2031 & 2973 \\
 & Mixtral & 1999 & 2932 \\ \midrule 
\multicolumn{4}{c}{Aggregated dataset} \\
IMDB & v2/v3/Mixtral & 4998 & 24797 \\
Emotion & v2/v3/Mixtral & 1978 & 1977 \\
Massive & v2/v3/Mixtral & 1998 & 2929 \\ \bottomrule
\end{tabular}
\vspace{-5mm}
\end{table}

\begin{table*}[!htb]
\caption{The results on the test sets of the IMDB, Emotion, and Amazon Massive datasets are presented. $T$ is the token temperature in the generation settings, and $\tau^{*}$ is the optimal label temperature tuned on the corresponding validation set. The ACC, ECE, and MCE are reported on a percentage scale (\%), this applies to other tables.}
\centering
\footnotesize
\setlength{\tabcolsep}{3.75pt}
\begin{tabular}{lcccccccccccccccc}
\toprule
 &  & \multicolumn{5}{c}{Claude-v2} & \multicolumn{5}{c}{Claude-v3} & \multicolumn{5}{c}{Mixtral} \\
\textbf{Methods} & \textbf{$T$} & \textbf{ACC} & \textbf{NLL} & \textbf{ECE} & \textbf{MCE} & \textbf{$\tau^{*}$} & \textbf{ACC} & \textbf{NLL} & \textbf{ECE} & \textbf{MCE} & \textbf{$\tau^{*}$} & \textbf{ACC} & \textbf{NLL} & \textbf{ECE} & \textbf{MCE} & \textbf{$\tau^{*}$} \\ \cmidrule(lr){3-7} \cmidrule(lr){8-12} \cmidrule(lr){13-17}
\multicolumn{17}{c}{IMDB} \\
Uncal. & 0.0 & 93.5 & 0.260 & 3.7 & 13.5 & /    & 94.2 & 0.200 & 5.2 & 45.0 & /    & 93.2 & 0.334 & 3.4 & 19.6 & /    \\
TS     & 0.0 & 93.5 & 0.244 & 3.1 & 13.6 & 1.12 & 94.2 & 0.185 & 1.0 & 52.5 & 0.64 & 93.2 & 0.276 & 1.8 & 23.4 & 1.39 \\
Uncal. & 1.0 & 93.1 & 0.298 & 3.2 & 19.5 & /    & 93.9 & 0.209 & 5.5 & 13.6 & /    & 91.6 & 0.383 & 4.0 & 39.2 & /    \\
TS     & 1.0 & 93.1 & 0.281 & 3.0 & 19.7 & 1.17 & 93.9 & 0.195 & 1.4 & 7.5  & 0.63 & 91.6 & 0.310 & 2.9 & 39.0 & 1.42 \\ \cmidrule(lr){3-7} \cmidrule(lr){8-12} \cmidrule(lr){13-17}
\multicolumn{17}{c}{Emotion Dataset} \\
Uncal. & 0.0 & 54.9 & 1.505 & 7.1 & 77.0 & /    & 54.0 & 2.106 & 7.8 & 80.0 & /    & 51.4 & 2.637 & 16.9 & 41.0 & /    \\
TS     & 0.0 & 54.9 & 1.450 & 3.3 & 26.5 & 1.26 & 54.0 & 2.301 & 3.9 & 33.1 & 0.83 & 51.4 & 1.979 & 4.9  & 33.3 & 1.75 \\
Uncal. & 1.0 & 53.2 & 1.874 & 8.8 & 31.6 & /    & 51.8 & 2.132 & 5.0 & 75.0 & /    & 52.2 & 2.591 & 15.4 & 27.8 & /    \\
TS     & 1.0 & 53.2 & 1.731 & 4.1 & 30.1 & 1.26 & 51.8 & 2.262 & 4.2 & 74.0 & 0.88 & 52.2 & 1.962 & 4.4  & 25.0 & 1.73 \\ \cmidrule(lr){3-7} \cmidrule(lr){8-12} \cmidrule(lr){13-17}
\multicolumn{17}{c}{Amazon Massive} \\
Uncal. & 0.0 & 73.6 & 3.205 & 14.5 & 58.9 & /    & 79.2 & 2.762 & 11.1 & 41.0 & /    & 68.9 & 3.374 & 15.3 & 38.7 & /    \\
TS     & 0.0 & 73.6 & 2.301 & 7.2  & 49.2 & 1.53 & 79.2 & 1.806 & 7.8  & 19.4 & 1.67 & 68.9 & 2.458 & 10.0 & 52.4 & 1.65 \\
Uncal. & 1.0 & 71.6 & 3.720 & 16.6 & 45.0 & /    & 78.5 & 2.912 & 11.1 & 75.0 & /    & 68.5 & 3.636 & 14.7 & 32.5 & /    \\
TS     & 1.0 & 71.6 & 2.534 & 9.4  & 23.7 & 1.60 & 78.5 & 2.28  & 7.8  & 16.6 & 1.33 & 68.5 & 2.612 & 9.9  & 54.2 & 1.64 \\ \bottomrule
\end{tabular}
\label{tab:result}
\end{table*}
\begin{figure}[!htb]
 \vspace{-2mm}
    \centering
    \subfloat[The percentage of sum of prob. to 1]{{ \includegraphics[width=0.375\textwidth]{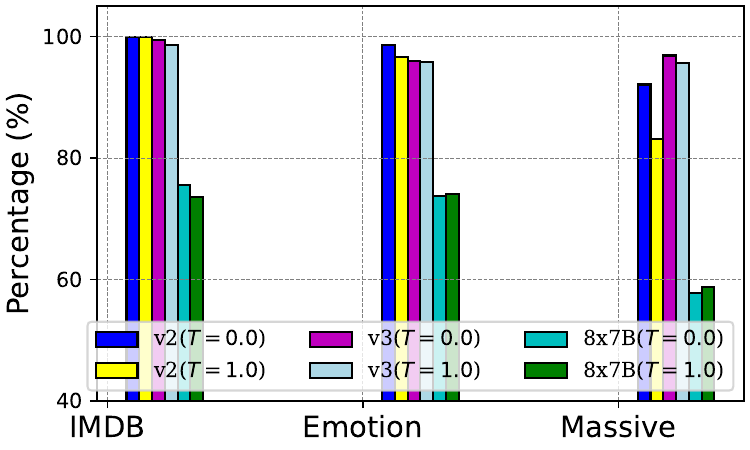}} } \vspace{5mm}
    \subfloat[The mean and variance of the sum of prob.]{{ \includegraphics[width=0.375\textwidth]{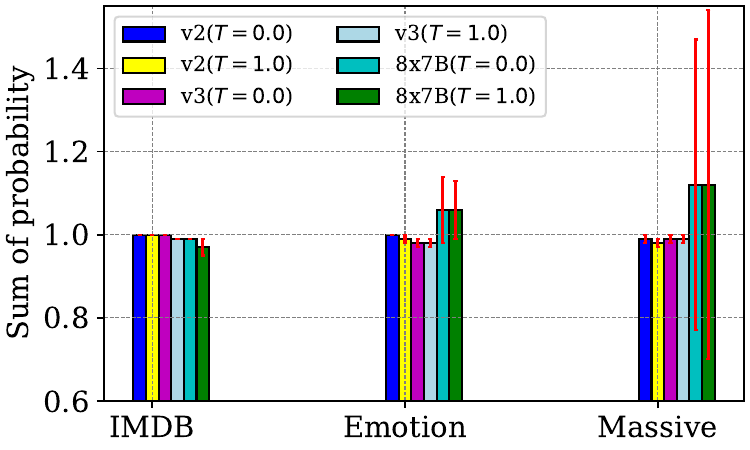}} }
    \caption{(a) The success rate of generated probability distribution and (b) the mean and variance of the sum of probability with different token temperatures $T$.}
    \label{fig:conf_sum_to_1_rate}
\end{figure}
\begin{figure*}[!htb]
\vspace{-4mm}
	\centering
 \subfloat[\scriptsize{Uncalibrated}]
    {{\includegraphics[width=0.2\textwidth]{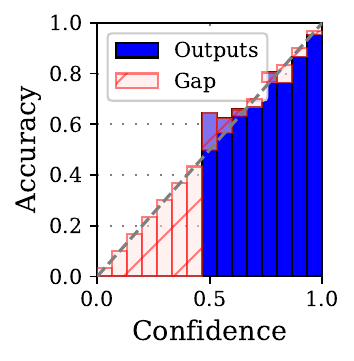} }}
    \subfloat[\scriptsize{Uncalibrated}]{{\includegraphics[width=0.2\textwidth]{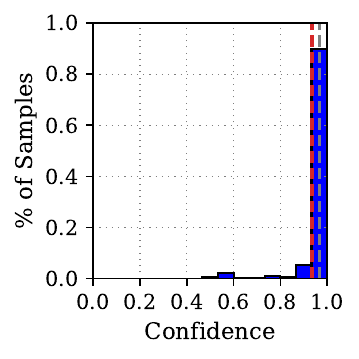} }}
    	\subfloat[\scriptsize{TS ($\tau$=1.12)}]
    {{\includegraphics[width=0.2\textwidth]{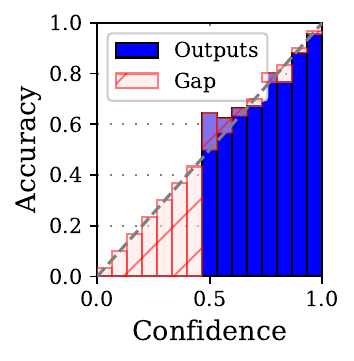} }}
    \subfloat[\scriptsize{TS ($\tau$=1.12)}]{{\includegraphics[width=0.2\textwidth]{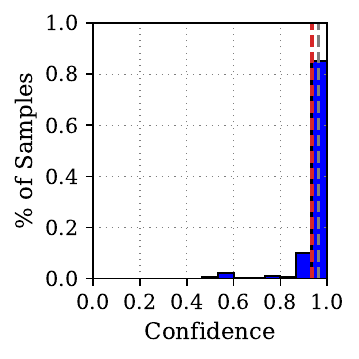} }}
 \\
  	\subfloat[\scriptsize{Uncalibrated}]
    {{\includegraphics[width=0.2\textwidth]{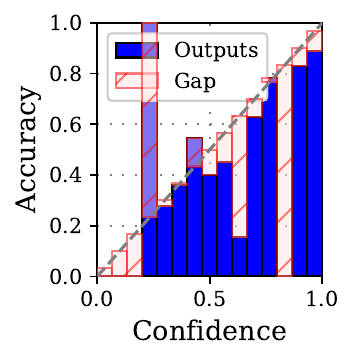} }}
    \subfloat[\scriptsize{Uncalibrated}]{{\includegraphics[width=0.2\textwidth]{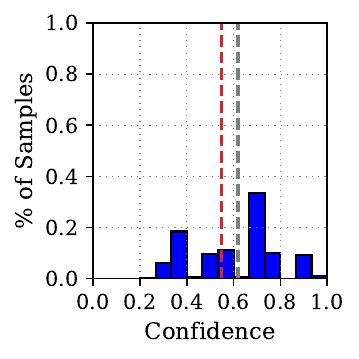} }}
    	\subfloat[\scriptsize{TS ($\tau$=1.26)}]
    {{\includegraphics[width=0.2\textwidth]{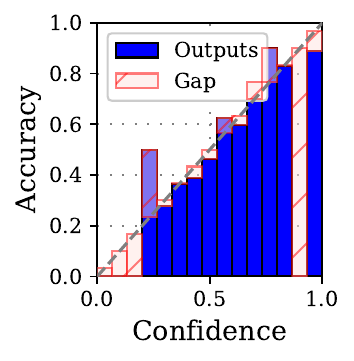} }}
    \subfloat[\scriptsize{TS ($\tau$=1.26)}]{{\includegraphics[width=0.2\textwidth]{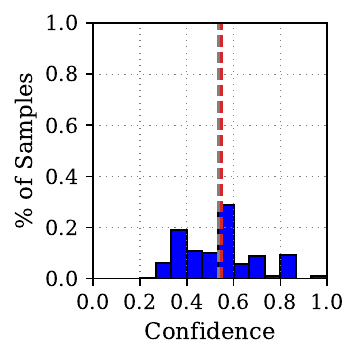} }} \\
 \subfloat[\scriptsize{Uncalibrated}]
    {{\includegraphics[width=0.2\textwidth]{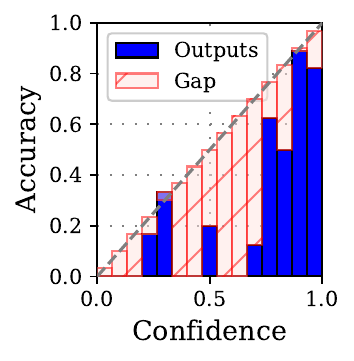} }}
    \subfloat[\scriptsize{Uncalibrated}]{{\includegraphics[width=0.2\textwidth]{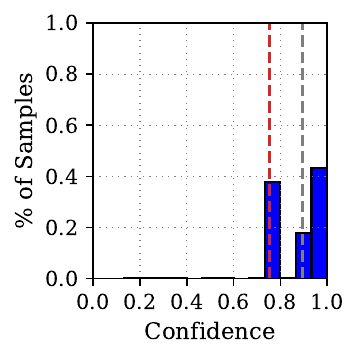} }}
    	\subfloat[\scriptsize{TS ($\tau$=1.53)}]
    {{\includegraphics[width=0.2\textwidth]{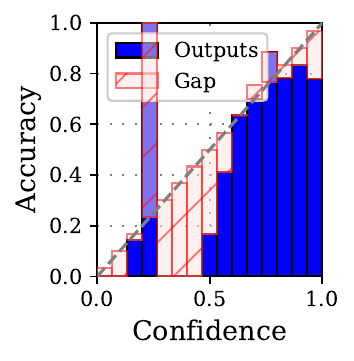} }}
    \subfloat[\scriptsize{TS ($\tau$=1.53)}]{{\includegraphics[width=0.2\textwidth]{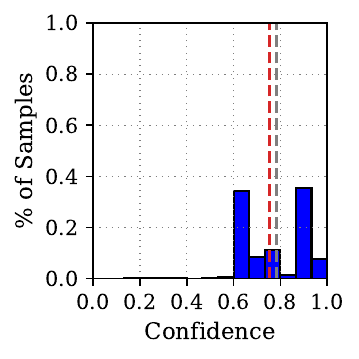} }} \\

	\caption{The reliability diagram and confidence histogram for the test set IMDB (top row), Emotion (middle row) and Amazon massive (bottom row). The $\tau$ is tuned based on TS with the corresponding validation set. It is clear to see that TS pushes the average confidence closer to accuracy. The result is based on the Claude-v2 with $t$=0.0. See section \ref{sec:add_res} in Appendix for the plots for Claude-v2 with $t$=1.0, Mixtral and Claude-v3.}
\label{fig:calibration}
\end{figure*}

\begin{table*}[htb]

\setlength{\tabcolsep}{3.75pt}
\footnotesize
\caption{The comparison of predictive and calibration performance on the aggregated test datasets (token temperature $T=0.0$. See Table~\ref{tab:agg_result_t1} in Appendix for $T=1.0$.}

\begin{tabular}{lccccccccccccccc}
\toprule
 & \multicolumn{5}{c}{Claude-v2} & \multicolumn{5}{c}{Claude-v3} & \multicolumn{5}{c}{Mixtral} \\ \cmidrule(lr){2-6} \cmidrule(lr){7-11} \cmidrule(lr){12-16}
\textbf{Datasets} & \textbf{ACC} & \textbf{NLL} & \textbf{ECE} & \textbf{MCE} & \textbf{$\tau^{*}$} & \textbf{ACC} & \textbf{NLL} & \textbf{ECE} & \textbf{MCE} & \textbf{$\tau^{*}$} & \textbf{ACC} & \textbf{NLL} & \textbf{ECE} & \textbf{MCE} & \textbf{$\tau^{*}$} \\
IMDB    & 93.5 & 0.244 & 3.1 & 13.6 & 1.12  & 94.2 & 0.184 & 1.0 & 52.5 & 0.64  & 93.2 & 0.276 & 1.8 & 23.4 & 1.39  \\
Emotion & 55.0 & 1.450 & 3.2 & 26.5 & 1.26  & 54.1 & 2.240 &   4.1 & 47.8 & 0.86  & 51.5 & 1.98  &  5.0 & 33.3 & 1.75  \\
Massive & 74.0 & 2.285 & 7.1 & 49.2 & 1.53  & 79.7 & 1.830 & 7.3 & 19.9 & 1.60  & 68.9 & 2.460 & 10.0 & 52.4 & 1.65  \\ \cmidrule(lr){2-6} \cmidrule(lr){7-11} \cmidrule(lr){12-16}
Average & 74.2 & 1.326 & 4.5 & 29.8 & 1.303 & 76.0 & 1.418 & 4.1 & 40.1 & 1.033 & 71.2 & 1.572 & 5.6 & 36.4 & 1.597 \\
\bottomrule
\end{tabular}
\label{tab:agg_result}
\end{table*}

 \begin{figure*}[!htb]
 \vspace{-3mm}
    \centering
    \subfloat[The probability difference]{{ \includegraphics[width=0.3\textwidth]{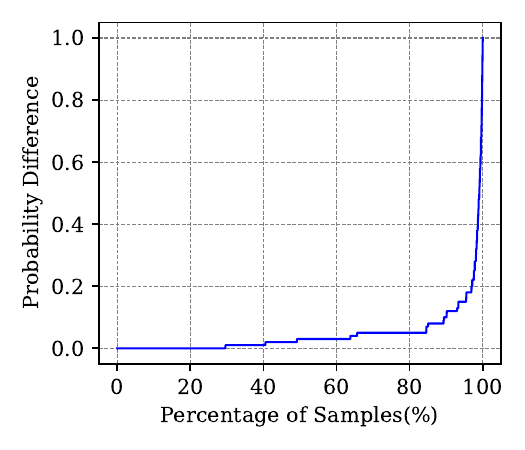}} } 
    \subfloat[The P-R curve for prob.(1 decimal)]{{ \includegraphics[width=0.3\textwidth]{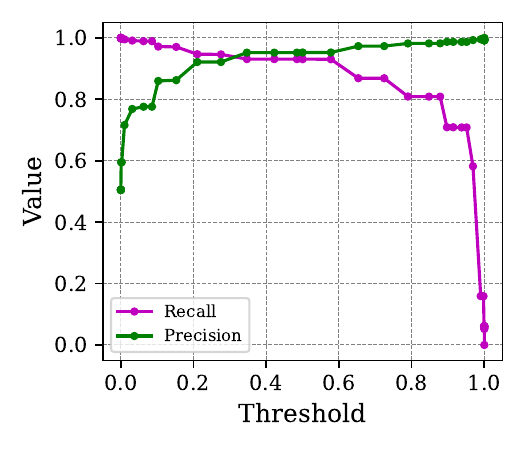}} } \vspace{5mm}
    \subfloat[The P-R curve for prob.(2 decimal)]{{ \includegraphics[width=0.3\textwidth]{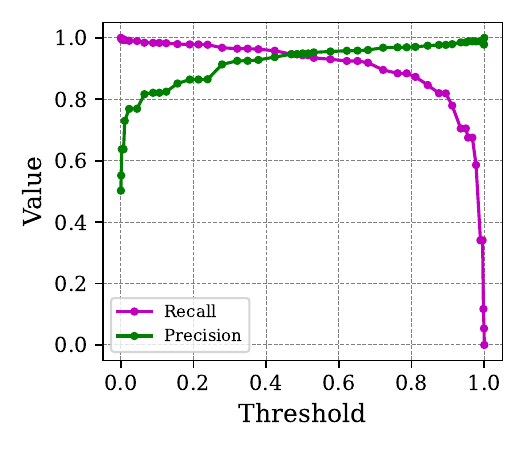}} }
\vspace{-3mm}
\caption{The comparison the probability with 1 decimal and 2 decimal on IMDB test set (positive class).}
\label{fig:cal_comp}
\end{figure*}

\subsection{Improved calibration with temperature scaling on the estimated logits}
Table \ref{tab:result} presents the zero-shot performance of Claude (v2, v3) and Mixtral on the test sets.  It demonstrates that calibrating logit proxy with TS improves calibration performance as compared to uncalibrated. This can be seen from the reduced ECE scores as well as the confidence histogram plots (the 2nd and 4th column) in Figure \ref{fig:calibration}. The TS reduces the gap between average confidence and accuracy. In generating probability distribution, we experimented with different token temperature ($T=0$ and $T=1$) in configuration. We observed that $T=0$ generally gives better accuracy as well as calibration scores across models and datasets.

\subsection{Black-box LLMs have strong capability in generating probability distribution}
\label{sec:gen_prob_dist}
To examine how LLMs understand confidence scores, which is supposed to ``softmax probability" and the capability of LLMs to generate probability distributions, we define a success criteria by computing the percentage of generating a probability distribution which sums to 1. 
\begin{equation}
    \textsc{success\_rate} = \frac{\mathbb{I} ({\sum_k^K p_k=1})}{N}
\end{equation}
where $N$ is the total number of instances. As presented in Figure~\ref{fig:conf_sum_to_1_rate} (a)Claude-v2 and v3 show a strong capability (>95\%) in generating correct probability distributions on the IMDB and Emotion datasets. This happens to a lesser extent on a harder dataset, the Amazon Massive dataset, which has 60 intent classes. We also noticed that using larger $T$ values tends to worsen the success rate on the Massive dataset, particularly for Claude-v2. One possible reason is that larger $T$ values generate tokens in a less deterministic way and the labels of the Massive data contain multiple tokens such as \texttt{cooking\_query}, \texttt{transport\_taxi} and \texttt{qa\_factoid}. On the other hand, we noticed that Mixtral gives lower success rate: 75.49\%, 73.8\%, 57.84\% on the three datasets respectively.  

Another aspect is to examine the quality of generated probability distribution, we measure the mean and variance of $p_{sum}$= $\sum_k^K p_k$, which ideally equals 1 if LLMs show correct understanding of confidence or softmax probability. Figure~\ref{fig:conf_sum_to_1_rate} (b) presents the mean and variance of $p_{sum}$. On IMDB, all LLMs achieve a mean close to 1 with low variance. From Emotion to Massive datasets, we observe increased variances. Particularly Mixtral shows high variances the two datasets.

\subsection{Claude outperforms Mixtral on both predictive and calibration performance}
To make a comparison across LLMs, we used the aggregated dataset as described in Table~\ref{tab:post_processed} (bottom). Table~\ref{tab:agg_result} presents both predictive and calibration performance. On IMDB datasets, the three LLMs gives similar accuracy, v3 achieves the best ECE score. On other two datasets, we see the performance drops of Claude-v2 and Mixtral, particularly, the latter has roughly 3.6\% and 10.8\% lower accuracy as compared to Claude-v3. In terms of calibration performance, Mixtral also shows worse performance to Claude-v2 and v3.

\subsection{Ablation study of generating probability with 2 decimal is beneficial}
\begin{table}[!htb]
\setlength{\tabcolsep}{4pt}
\centering
\small
\caption{Performance comparison between probability with 1 and 2 decimal digits. The results are based on Claude-v3 on IMDB test set.}
\begin{tabular}{lccccc}
\toprule
\textbf{} & \textbf{ACC} & \textbf{NLL} & \textbf{ECE} & \textbf{MCE} & \textbf{$\tau^{*}$} \\
Prob. (1 decimal) & 94.2 & 0.184 & 1.0 & 52.5 & 0.64 \\
Prob. (2 decimal) & 94.6 & 0.187 & 1.8 & 10.2 & 0.65 \\
\bottomrule
\end{tabular}
\label{tab:decimal_com}
\vspace{-4mm}
\end{table}
One of observed limitations of verbalised probability is that, without more specific instruction, LLMs generate less diversely distributed probability, mostly around the 11 different probability scores(i.e., \{0.0, 0.1, ..., 1.0\}). This would lead to an issue that many different samples have same probability score. One of most useful applications of calibrated probability is to perform precision-recall threshold tuning and control model towards expected precision or recall. When many samples have same probability, this actually prevents a calibrated probability be further used in a more controllable setup. Therefore, we conducted a group of experiment on IMDB with Claude-v3 to generate more fine-grained probability i.e., with 2 decimal.

Table~\ref{tab:decimal_com} presents the performance comparison. We can observe that, in general, the generated probability with 2 decimal is beneficial. Though it gives bit worse NLL and ECE, MCE score is largely improved. The Figure~\ref{fig:cal_comp} (a) reports the probability difference, >95\% samples have probability difference <0.2. However, we can clearly see better distributed probability of using 2 decimal, which leads to smoother Precision-Recall curve as shown in Figure~\ref{fig:cal_comp} (b) and (c)).

On the other hand, in some cases, LLM predictions and verbalized probability can be noisy; the nuance of requesting LLM to respond probability with 2 digits or 1 digit can actually flip the label outcome and cause a large variance (we found 2 such examples from test set as shown in Appendix ~\ref{tab:examples_diff}).
Qualitatively, these examples which flip may be more nuanced in terms of language use (i.e. potentially harder to predict).


\section{Related Work}


\textbf{Calibration with accessible probabilities}.
\citet{guo2017calibration} and \citet{mukhoti2020calibrating} showed that the miscalibration of large modern neural networks (NNs) is related to overfitting on the likelihood of the training dataset. To tackle this, many methods have been proposed such as focal loss~\citep{lin2017focal}, which acts as a maximum entropy regularizer \cite{mukhoti2020calibrating}. Label smoothing~\cite{muller2019does}, Bayesian approach~\citep{maddox2019simple}, meta-learning~\citep{bohdal2021meta}, the Gumbel-softmax Trick~\citep{jang2017categorical,wang2020uncertainty,pei2022transformer}, and kernel-based methods~\citep{kumar2018trainable}. The recent survey summarizes these approaches~\cite{wang2023calibration}.

\textbf{Calibration with inaccessible probabilities}. 
For closed-source LLMs, the aforementioned methods are not applicable. To address this limitation, LLMs are prompted to generate verbalized confidence over their outputs~\cite{lin2022teaching}. This technique was introduced by \citet{tian-etal-2023-just} to infer probabilities in Question-Answering (QA) tasks, demonstrating better calibration than conditional probabilities. \citet{rivera2024combining} integrated this approach with sampling methods for uncertainty quantification. \citet{xiong2023can} further extended it by combining it with sampling strategies to generate multiple responses and employing an aggregation approach for response consistency.

Another line of work from \citet{ulmer2024calibrating} focuses on training an auxiliary model to calibrate verbalized probabilities. Upon obtaining confidence scores from generation, the authors employ Platt scaling~\citep{Platt1999,guo2017calibration} to adjust these scores. Conceptually different to those work, we focus on prompting LLMs to generate categorical distribution and employing temperature scaling on estimated logits for post-hoc calibration.

\section{Discussion and Conclusion}
We demonstrated that LLMs are able to generate probability distribution, which offers access to model verbalized confidence. We then identified the re-softmax issue on verbalized probability on discriminative tasks. We demonstrated that performing temperature scaling on reverted verbalized probability can effectively improve LLMs predictive calibration. This is achieved by leveraging revert softmax trick. The empirical results on three public datasets and one internal dataset evidently demonstrate the effectiveness of our proposed method.

We also showed the beneficial of generated probability with 2 decimal, and revealed the possible high variances for some examples.
Measuring and turning this variance into an asset in terms of assigned label and probability in order to further improve robustness of LLM labels and calibration of verbalized probabilities is future work.

\section*{Limitations}

As mentioned in Sec.~\ref{sec:experiments}, one of the limitations of the current approach is that black-box LLMs can fail to precisely follow instructions and output something other than confidence scores, even when explicitly asked to do so. This forces future users to either tune the prompts to diminish this behavior as much as possible, or to implement additional postprocessing logic to extract the actual scores. However, even when doing so there are no guarantees that the LLM will return well-formed outputs for every input.

\bibliography{custom}
\bibliographystyle{acl_natbib}

\newpage
\appendix

\onecolumn
\section{Appendix}
\label{sec:appendix}
\subsection{Claude v2 and Mixtral Prompt Templates}
\label{sec:appendix:prompts-1}

Here we show the prompts we used with Claude v2 and Mixtral in the experiments described in \cref{sec:experiments}.

\label{table:claude-v2-prompts}
\noindent \textbf{IMDB}

\begin{tcolorbox}
\begin{lstlisting}[basicstyle=\ttfamily\footnotesize, breaklines=true, breakatwhitespace=true]
Give a binary sentiment label (positive or negative) to the following sentence: \$text.

Assign a confidence score (between 0 and 1.0) to this prediction.

Give ONLY the probability distribution over the sentiment labels.

Give ONLY the probability, no other words or explanation.

Provide ONLY the probability in a format of Python dict.
\end{lstlisting}
\end{tcolorbox}

\noindent \textbf{Emotion}

\begin{tcolorbox}
\begin{lstlisting}[basicstyle=\ttfamily\footnotesize, breaklines=true, breakatwhitespace=true]
Give an emotion label from a label list [sadness, joy, love, anger, fear, surprise] to the following sentence: \$text.

Assign a confidence score (between 0 and 1.0) to this prediction.

Give ONLY the probability distribution over the 6 emotion labels.

Give ONLY the probability, no other words or explanation.

Provide ONLY the probability in a format of Python dict.
\end{lstlisting}
\end{tcolorbox}

\noindent \textbf{Amazon Massive}

\begin{tcolorbox}
\begin{lstlisting}[basicstyle=\ttfamily\footnotesize]
Give a label from the following 60 intent labels:
["datetime_query", "iot_hue_lightchange", "transport_ticket",
 "takeaway_query", "qa_stock", "general_greet",
 "recommendation_events", "music_dislikeness", "iot_wemo_off",
 "cooking_recipe", "qa_currency", "transport_traffic",
 "general_quirky", "weather_query", "audio_volume_up",
 "email_addcontact", "takeaway_order", "email_querycontact",
 "iot_hue_lightup", "recommendation_locations",
 "play_audiobook", "lists_createoradd", "news_query",
 "alarm_query", "iot_wemo_on", "general_joke", "qa_definition",
 "social_query", "music_settings", "audio_volume_other",
 "calendar_remove", "iot_hue_lightdim", "calendar_query",
 "email_sendemail", "iot_cleaning", "audio_volume_down",
 "play_radio", "cooking_query", "datetime_convert", "qa_maths",
 "iot_hue_lightoff", "iot_hue_lighton", "transport_query",
 "music_likeness", "email_query", "play_music",
 "audio_volume_mute", "social_post", "alarm_set", "qa_factoid",
 "calendar_set", "play_game", "alarm_remove", "lists_remove",
 "transport_taxi", "recommendation_movies", "iot_coffee",
 "music_query", "play_podcasts", "lists_query"]
to the following sentence: $text.
\end{lstlisting}
\end{tcolorbox}

\newpage

\subsection{Claude v3 Prompt Templates}
\label{sec:appendix:prompts-2}

The prompts we used with Claude v3 in the experiments described in \cref{sec:experiments} are shown below.

\noindent \textbf{IMDB}
\begin{tcolorbox}
\begin{lstlisting}[basicstyle=\ttfamily\footnotesize, breaklines=true, breakatwhitespace=true]
{
  "system": "Give a binary sentiment label (negative or positive) for a given text.\nAssign a confidence score (between 0 and 1.0) to this prediction.\nGive ONLY the probability distribution over the 2 sentiment labels.\nGive ONLY the probability, no other words or explanation.\n Provide ONLY the probability in a format of Python dict.",
  "messages": [{"role": "user", "content": "$text" }]
}
\end{lstlisting}
\end{tcolorbox}

\noindent \textbf{Emotion}
\begin{tcolorbox}
\begin{lstlisting}[basicstyle=\ttfamily\footnotesize, breaklines=true, breakatwhitespace=true]
{
  "system": "Give an emotion label from a label list [sadness, joy, love, anger, fear, surprise] for a given text.\nAssign a confidence score (between 0 and 1.0) to this prediction.\nGive ONLY the probability distribution over the 6 emotion labels.\nGive ONLY the probability, no other words or explanation.\n Provide ONLY the probability in a format of Python dict.",
  "messages": [{"role": "user", "content": "$text" }]
}
\end{lstlisting}
\end{tcolorbox}

\noindent \textbf{Amazon Massive}
\begin{tcolorbox}
\begin{lstlisting}[basicstyle=\ttfamily\footnotesize, breaklines=true, breakatwhitespace=true]
{
  "system": "Give a label from the following 60 intent labels ['datetime_query','iot_hue_lightchange','transport_ticket','takeaway_query', 'qa_stock','general_greet','recommendation_events','music_dislikeness', 'iot_wemo_off','cooking_recipe','qa_currency','transport_traffic', 'general_quirky','weather_query','audio_volume_up','email_addcontact', 'takeaway_order', 'email_querycontact','iot_hue_lightup', 'recommendation_locations', 'play_audiobook', 'lists_createoradd', 'news_query', 'alarm_query','iot_wemo_on', 'general_joke', 'qa_definition', 'social_query', 'music_settings', 'audio_volume_other','calendar_remove', 'iot_hue_lightdim', 'calendar_query', 'email_sendemail', 'iot_cleaning', 'audio_volume_down','play_radio', 'cooking_query', 'datetime_convert', 'qa_maths', 'iot_hue_lightoff', 'iot_hue_lighton', 'transport_query', 'music_likeness', 'email_query', 'play_music', 'audio_volume_mute', 'social_post', 'alarm_set', 'qa_factoid', 'calendar_set','play_game', 'alarm_remove', 'lists_remove', 'transport_taxi', 'recommendation_movies', 'iot_coffee', 'music_query','play_podcasts', 'lists_query'] for a given text.\nAssign a confidence score (between 0 and 1.0) to this prediction.\nGive ONLY the probability distribution over the 60 intent labels.\nGive ONLY the probability, no other words or explanation.\n Provide ONLY the probability in a format of Python dict.",
  "messages": [{"role": "user", "content": "$text" }]
}
\end{lstlisting}
\end{tcolorbox}

\newpage

\subsection{Proofs}
\label{proof:prop_1}
\noindent \textbf{Proposition 2}
\textit{Given a categorical probability distribution $\bfp = \{p_1,...,p_i,...,p_K\}$ over $K$ ($K>1$) classes, and a re-softmaxed probability distribution $\bfq = \textsc{softmax}(\bfp) = \{q_1,...,q_i,...,q_K\}$, for $i\in [1, K]$, the $q_i$ is bounded: $0<\frac{1}{K-1+e} \leq q_i \leq \frac{e}{K-1+e}< 1$}

\begin{proof}
For probability distribution $\bfp = \{p_1,...,p_i,...,p_K\}$ over $K$($K>1$) classes, we have $0 \leq p_i \leq 1$ and $\sum_i^K p_i=1$. Then we have $1 \leq e^{p_i} \leq e $. Re-softmaxing $\bfp$, gives 
\begin{align}
    q_i &= \frac{e^{p_i}}{\sum_{j=1}^K e^{p_j}} \\
        &= \frac{e^{p_i}}{e^{p_1} + e^{p_2} + ... e^{p_j} + e^{p_K}}
\end{align}
Assume $p_i=1$ (the upper bound of in original probability $\bfp$), then $p_j=0$, for $j \in [1, K]$ and $i\neq j$, and $e^{p_i}=e, e^{p_j}=1$ rewrite eq.(8), we can find the upper bound for $q_i$:
\begin{align}
    q_i & \leq \frac{e}{1 + 1 + e + 1 + ... + 1} \\
    & \leq \frac{e}{K-1+e} 
\end{align}
\end{proof}
To find the lower bound of $q_i$, we can assume $p_i=0$ (the lower bound of in original probability $\bfp$), then we rewrite the eq.(8) to 
\begin{align}
    q_i &= \frac{1}{e^{p_1} + e^{p_2} + 1 + ... e^{p_j} + e^{p_K}},  ~~~~j \in [1, K], i\neq j\\
    & \geq  \frac{1}{1 + \max(e^{p_1} + e^{p_2} + ... e^{p_j} + e^{p_K})} \\
    & \geq  \frac{1}{1 + (1 + 1 + ... e + 1)} \\
    & \geq  \frac{1}{1 + (K - 2 + e)} \\
    & \geq  \frac{1}{K -1 + e} 
\end{align}
Note that in eq.(12), when $p_i=0$, the maximum given by $\max(e^{p_1} + e^{p_2} + ... e^{p_j} + e^{p_K})$ is when $p_j=1,i\neq j $ and the rest of $K-2$ class have probability 0.

For examples: 
$\bfp$=$[0, 1]$, $\bfq$=$[\frac{1}{2-1+e}, \frac{e}{2-1+e}]= [0.2689, 0.7311]$;

$\bfp$=$[0,0,0,0,1]$, $\bfq$=$[\frac{1}{5-1+e}, \frac{1}{5-1+e},\frac{1}{5-1+e},\frac{1}{5-1+e},\frac{e}{5-1+e}]$ =$[0.14885, 0.14885, 0.14885, 0.14885, 0.40460]$ 

For classification tasks, the minimum number of classes $K=2$ (i.e., binary classification), thus we have $0 < \frac{1}{K -1 + e} \leq q_i \leq \frac{e}{K -1 + e} < 1$.
~\\ \newpage

\noindent \textbf{Proposition 2}
\textit{In the context of calibrating verbalized probability using temperature scaling (TS), let \(\tau_p\) and \(\tau_q\) represent the optimal temperature values for calibrating \(\bfp\) and \(\bfq\) respectively. It follows that \(\tau_q < \tau_p\)}.
\begin{proof}
\label{proof_prop_2}
Let $\bfc$ be the true probability, assume TS calibrated probability is $\bfp^{*}$ from calibrating $\bfp$: $\bfp^{*} = \textsc{Softmax}(\frac{\bfp}{\tau_p})$, and TS calibrated probability is $\bfq^{*}$ from calibrating $\bfq$: $\bfq^{*} = \textsc{Softmax}(\frac{\bfq}{\tau_q})$. On the same dataset, for a perfectly calibrated probability have $\bfc=\bfp^{*}=\bfq^{*}$. Then we have:
\begin{equation}
    \textsc{softmax} (\frac{\bfp}{\tau_p}) = \textsc{softmax} (\frac{\bfq}{\tau_q})
\end{equation}
from Proposition 1, we know that $\bfq$ has more uniform distribution than $\bfp$, i.e., entropy $\mathcal{H}(\bfq)$ > $\mathcal{H}(\bfp)$, predicative probability $\bfp$ is more confident. This requires $\tau_q$ < $\tau_p$.
\end{proof}

\subsection{Invert softmax implementation}
\label{sec:invert_softmax_imp}
This is the Python implementation of the invert softmax function that we used in our experiments:
\begin{python}
import numpy as np

def invert_softmax_prob(probs) -> np.array:
    """
    The inverse softmax function.
    probs: array-like, a vector of the softmax probabilities.
    returns: ndarray of estimated unnormalised logits, shape [n_samples, n_classes].
    """
    # offset with a small number to avoid division by zero
    logp = np.log(np.array(probs) + 1e-9)
    n_sample, n_class = logp.shape[0], logp.shape[1]
    # c can be also fixed to be 0.0 or 1.0
    c = -np.sum(logp, axis=1) / n_class
    # broadcasting c over n_samples
    offsets = np.repeat(c, n_class).reshape(n_sample, -1)
    logits = logp + offsets
    return logits
\end{python}

\begin{table}[!ht]
\centering
\caption{Examples of flipped prediction for using the probability (of being positive) with 1 and 2 decimals.}
\label{tab:examples_diff}
\begin{tabular}{p{7cm}cccc}
\hline
Input Text & True Label & $p_{1d}(1|x)$ & $p_{2d}(1|x)$ &   $\left |p_{1d}(1|x) - p_{2d}(1|x)\right |$ \\ \hline
This is not so much of a review as it is a testament that it has been proven, yet again, that the Academy rewards money, not artistic accomplishment. And I must say I am saddened that this usually artistic and intelligent band of imbd members have left this off the top 250. Boogie Nights is powerful, raw, and gutsy through script, direction and acting. Very few movies can claim this triple crown. & 1 & 0,9 & 0,15 & 0,75 \\ \hline
Aya! If you are looking for special effects that are 10-20 years before its time, this is it. The glowing lightning bolts, fireballs, etc. look like they came from a cheesy 70's sci-fi flick. And yes, Hercules really grows; he's not being pushed on a cart closer to the camera! & 0 & 0,9 & 0,15 & 0,75 \\ \hline
\end{tabular}
\end{table}

\subsection{Additional Results on The Aggregated Test Datasets}

\begin{table*}[htb]

\setlength{\tabcolsep}{3.75pt}
\footnotesize
\caption{Comparison of predictive and calibration performance on the aggregated test datasets (token temperature $T=1.0$).}
\begin{tabular}{lccccccccccccccc}
\toprule
 & \multicolumn{5}{c}{Claude V2} & \multicolumn{5}{c}{Claude V3} & \multicolumn{5}{c}{Mixtral} \\ \cmidrule(lr){2-6} \cmidrule(lr){7-11} \cmidrule(lr){12-16}
\textbf{Datasets} & \textbf{ACC} & \textbf{NLL} & \textbf{ECE} & \textbf{MCE} & \textbf{$\tau^{*}$} & \textbf{ACC} & \textbf{NLL} & \textbf{ECE} & \textbf{MCE} & \textbf{$\tau^{*}$} & \textbf{ACC} & \textbf{NLL} & \textbf{ECE} & \textbf{MCE} & \textbf{$\tau^{*}$} \\
IMDB    & 93.1 & 0.281 & 3.0 & 19.7 & 1.17  & 93.9 & 0.195 & 1.4 & 7.3 & 0.63  &  91.5 & 0.310 & 2.9 & 39.0 & 1.42  \\
Emotion & 53.2 & 1.733 & 3.9 & 30.1 & 1.26  & 52.0 & 2.273 & 4.2 & 74.0 & 0.88  & 51.5 & 1.98  & 5.0 & 33.3 & 1.75  \\
Massive & 71.9 & 2.790 & 9.7 & 29.4 & 1.39  & 78.9 & 2.321 & 8.0 & 19.7 & 1.28  & 68.5 & 2.611 & 9.9 & 54.2 & 1.64  \\ \cmidrule(lr){2-6} \cmidrule(lr){7-11} \cmidrule(lr){12-16}
Average & 72.7 & 1.601 & 5.5 & 26.4 & 1.273 & 74.9 & 1.596 & 4.5 & 33.7 & 0.930 & 70.5 & 1.634 & 5.9 & 42.2 & 1.603 \\
\bottomrule
\end{tabular}
\label{tab:agg_result_t1}
\end{table*}

\begin{table*}[htb]

\setlength{\tabcolsep}{3.75pt}
\footnotesize
\caption{Comparison of predictive and calibration performance on the aggregated test datasets with temperature scaling directly on generated verbalised probability, i.e., calibration with re-softmaxed probability (token temperature $T=0.0$). }
\begin{tabular}{lccccccccccccccc}
\toprule
 & \multicolumn{5}{c}{Claude-v2} & \multicolumn{5}{c}{Claude-v3} & \multicolumn{5}{c}{Mixtral} \\ \cmidrule(lr){2-6} \cmidrule(lr){7-11} \cmidrule(lr){12-16}
\textbf{Datasets} & \textbf{ACC} & \textbf{NLL} & \textbf{ECE} & \textbf{MCE} & \textbf{$\tau^{*}$} & \textbf{ACC} & \textbf{NLL} & \textbf{ECE} & \textbf{MCE} & \textbf{$\tau^{*}$} & \textbf{ACC} & \textbf{NLL} & \textbf{ECE} & \textbf{MCE} & \textbf{$\tau^{*}$} \\
IMDB & 93.5 & 0.224 & 1.6 & 15.9 & 0.30 & 94.2 & 0.175 & 0.8 & 59.6 & 0.22 & 93.2 & 0.213 & 2.4 & 47.5 & 0.31 \\
Emotion & 55.0 & 1.330 & 3.4 & 20.5 & 0.29 & 54.1 & 1.319 & 4.8 & 25.5 & 0.22 & 51.5 & 1.354 & 8.8 & 21.1 & 0.37 \\
Massive & 74.0 & 1.539 & 5.0 & 45.8 & 0.17 & 79.7 & 1.185 & 2.2 & 24.3 & 0.16 & 68.9 & 1.738 & 6.5 & 41.4 & 0.17 \\  \cmidrule(lr){2-6} \cmidrule(lr){7-11} \cmidrule(lr){12-16}
Average & 74.2 & 1.031  & 3.3 & 27.4 & 0.25  & 76.0 & 0.893 & 2.6 & 36.5 & 0.20 & 71.2 & 1.102 & 5.9 & 36.7 & 0.28 \\
\bottomrule
\end{tabular}
\label{tab:agg_result_t0_with_resoftmax}
\end{table*}

\subsection{Additional Results for Ablation Study}
\begin{figure}
	\centering
   \includegraphics[width=0.5\textwidth]{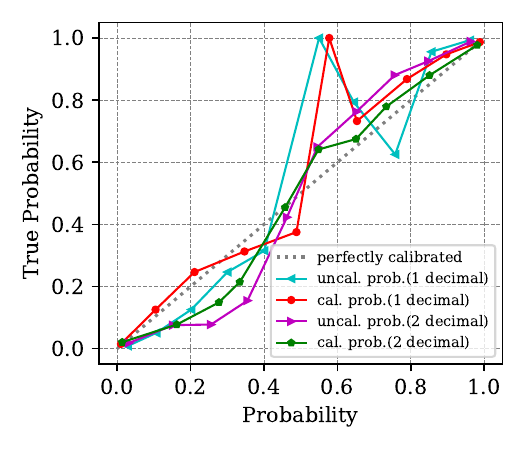}
   \caption{The calibration curve of Claude v3 on IMDB test set for the generated probability with 1 decimal and 2 decimal.}
   \label{fig:cal_curve_1d_2d}
  \end{figure}

Figure~\ref{fig:cal_curve_1d_2d} presents the comparison between the generated probability with 1 decimal and 2 decimal. We can see that the calibrated probabilities improves uncalibrated ones for both 1 decimal and 2 decimal; The 2 decimal-based probability improves 1 decimal-based probability. 
  
Table~\ref{tab:examples_diff} presents the examples that LLMs flipped prediction for a given text input by just request a probability with 2 decimal. The variance is 0.75. 

Table~\ref{tab:agg_result_t1} presents the comparison of predictive and calibration performance on the aggregated test datasets (token temperature $T=1.0$). 

Table~\ref{tab:agg_result_t0_with_resoftmax} reports the comparison of predictive and calibration performance on the aggregated test datasets with temperature scaling directly on generated verbalized probability, i.e., calibration with re-softmaxed probability (token temperature $T=0.0$). Comparing the above table to Table~\ref{tab:agg_result} in the paper, we can observe that directly applying temperature scaling to verbalized probabilities results in slightly lower calibration errors. However, this does not mean that directly scaling verbalized probabilities is better because it is not a correct calibration procedure. This procedure triggers re-softmaxing, which first pushes the probability distribution to be more flattened (under-confident) than the original distribution (Proposition~\ref{proposition_1}). To achieve calibrated probabilities that reflect true probabilities, temperature scaling then has to use a small temperature value to make the distribution reach the true confidence level (Proposition~\ref{proposition_2}). That's why we see significantly lower optimal temperature values ($\tau^*$) in the above table compared to the temperature values in Table ~\ref{tab:agg_result}, for example, average temperature on Mxitral 0.28 (temperature scaling on verbalized probability  ) vs 1.597 (temperature scaling on estimated logits from verbalized probability).  

As pointed out by recent research works~\cite{guo2017calibration, minderer2021revisiting}, modern neural networks (including large vision models or LLMs, those are mainly trained with standard softmax layer where temperature can be treated as 1.0) are primarily overconfident. This typically requires a temperature value $\tau^* > 1$ to make the models less overconfident, and $\tau^* < 1$ to make the models less underconfident. Conversely, directly applying temperature scaling to verbalized probabilities contradicts that theory, suggesting $\tau^* < \frac{e}{K-1+e} < 1$ ( $K$ is the number of classes) to make LLMs less overconfident.

\subsection{Ablation Experiment on Label Positional Bias}
We conducted a set of experiments to examine the sensitivity of the label order in the prompt with the Emotion dataset (6 classification labels)~\cite{saravia-etal-2018-carer}. We performed experiments with the exact same setup as described in Section~\ref{sec:experiments}, except for changing the label order. There are 5 runs: 1× the default label order (as in the paper); 1× the reversed default label order; 3× randomly shuffled label orders. 

Table~\ref{tab:label_positional_bias} presents the ablation comparison between a single run and multiple runs on the Emotion dataset with Claude-v3 (Token temperature = 0.0). Here we report the mean and variance across the 5 runs. We also include the corresponding single run result from Table~\ref{tab:result} for comparison. We do see that this positional bias leads to slightly different results. However, this difference is modest, and the metric numbers reported do not drift much from the mean numbers.

\begin{table}[!htb]
\setlength{\tabcolsep}{4pt}
\centering
\small
\caption{The results of examining label positional bias in prompt. The results are based on Claude-v3 on Emotion test set.}
\begin{tabular}{lccccc}
\toprule
\textbf{Method} & \textbf{ACC} & \textbf{NLL} & \textbf{ECE} & \textbf{MCE} & \textbf{$\tau^{*}$} \\
Single run (copied from Table 3) & 0.54 & 2.301 & 0.039 & 0.331 & 0.83 \\
Averaged results on 5 runs  & 0.542 $\pm$ 0.004 & 2.581 $\pm$ 0.05 & 0.038 $\pm$ 0.001 & 0.313 $\pm$ 0.027 & 0.958 $\pm$ 0.015 \\ 
\bottomrule
\end{tabular}
\label{tab:label_positional_bias}
\vspace{-4mm}
\end{table}

\subsection{Additional Plots for Reliability Diagrams and Confidence Histograms}
\label{sec:add_res}
\begin{figure}[htbp]
	\centering
	\subfloat[\scriptsize{Uncalibrated}]{{\includegraphics[width=0.2\textwidth]{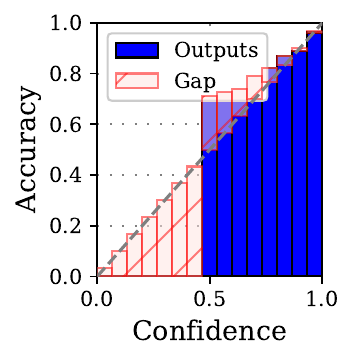} }}
	\subfloat[\scriptsize{Uncalibrated}]{{\includegraphics[width=0.2\textwidth]{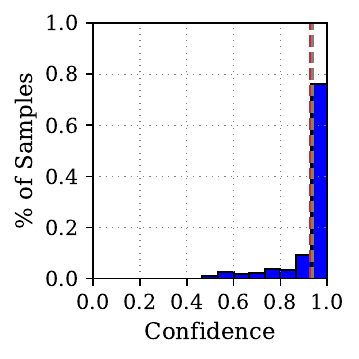} }}
	\subfloat[\scriptsize{TS ($\tau=1.17$)}]{{\includegraphics[width=0.2\textwidth]{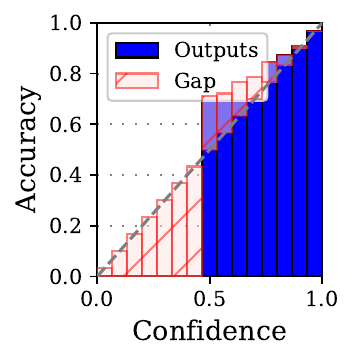} }}
	\subfloat[\scriptsize{TS ($\tau=1.17$)}]{{\includegraphics[width=0.2\textwidth]{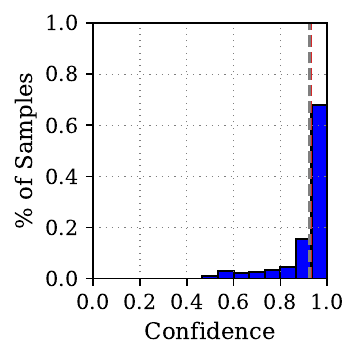} }} \\
	\subfloat[\scriptsize{Uncalibrated}]{{\includegraphics[width=0.2\textwidth]{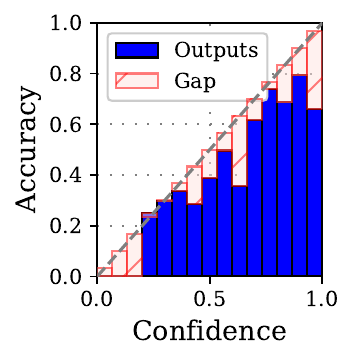} }}
	\subfloat[\scriptsize{Uncalibrated}]{{\includegraphics[width=0.2\textwidth]{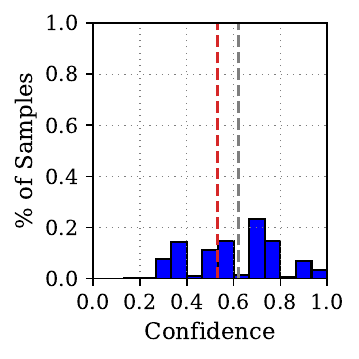} }}
	\subfloat[\scriptsize{TS ($\tau=1.26$)}]{{\includegraphics[width=0.2\textwidth]{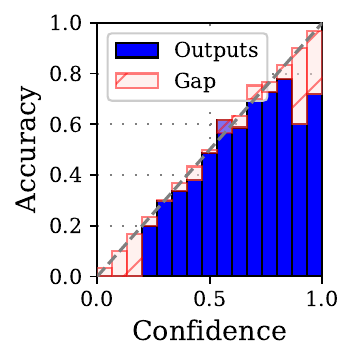} }}
    \subfloat[\scriptsize{TS ($\tau=1.26$)}]{{\includegraphics[width=0.2\textwidth]{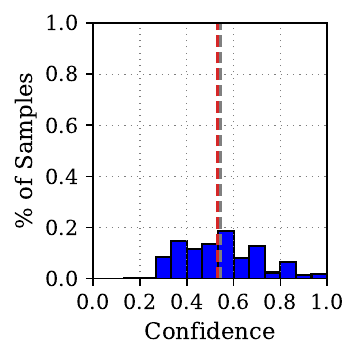} }} \\
	\subfloat[\scriptsize{Uncalibrated}]{{\includegraphics[width=0.2\textwidth]{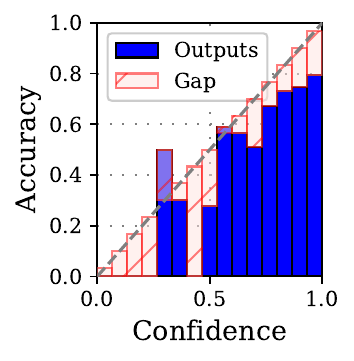} }}
	\subfloat[\scriptsize{Uncalibrated}]{{\includegraphics[width=0.2\textwidth]{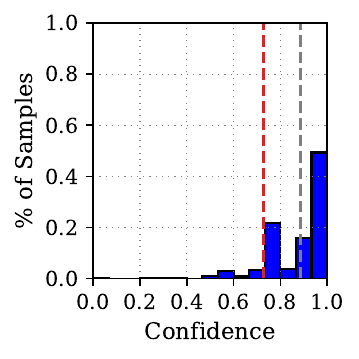} }}
	\subfloat[\scriptsize{TS ($\tau=1.59$)}]{{\includegraphics[width=0.2\textwidth]{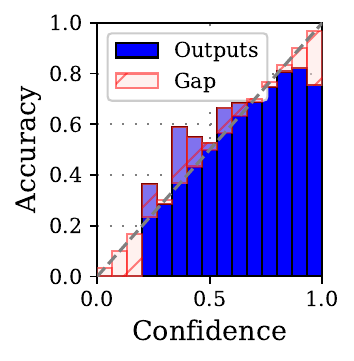} }}
	\subfloat[\scriptsize{TS ($\tau=1.59$)}]{{\includegraphics[width=0.2\textwidth]{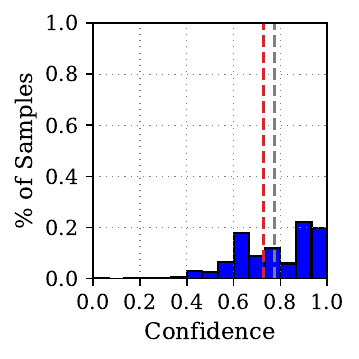} }} \\

	\caption{Claude v2 ($T=1.0$) reliability diagrams and confidence histograms for the test splits of the IMDB (top row), Emotion (middle row) and Amazon massive (bottom row) datasets. $\tau$ is tuned with TS on each corresponding validation dataset.}
\label{fig:calibration-claude-v2-T1}
\end{figure}

\begin{figure}[htbp]
	\centering
	\subfloat[\scriptsize{Uncalibrated}]{{\includegraphics[width=0.2\textwidth]{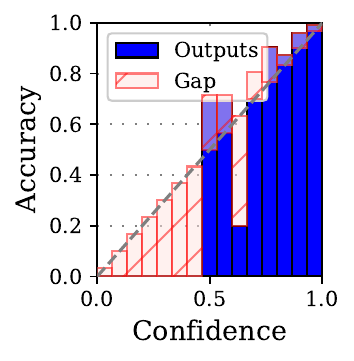} }}
	\subfloat[\scriptsize{Uncalibrated }]{{\includegraphics[width=0.2\textwidth]{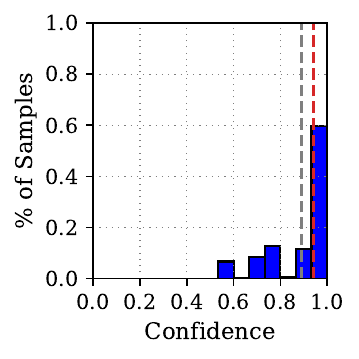} }}
	\subfloat[\scriptsize{TS ($\tau=0.64$)}]{{\includegraphics[width=0.2\textwidth]{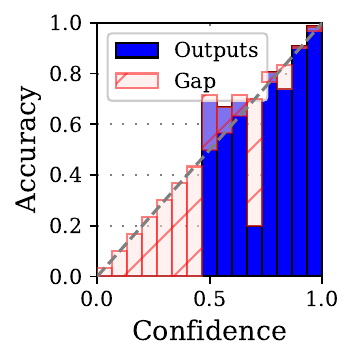} }}
	\subfloat[\scriptsize{TS ($\tau=0.64$)}]{{\includegraphics[width=0.2\textwidth]{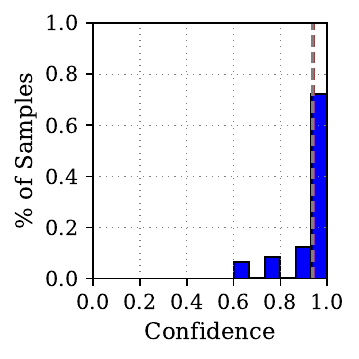} }} \\
	\subfloat[\scriptsize{Uncalibrated}]{{\includegraphics[width=0.2\textwidth]{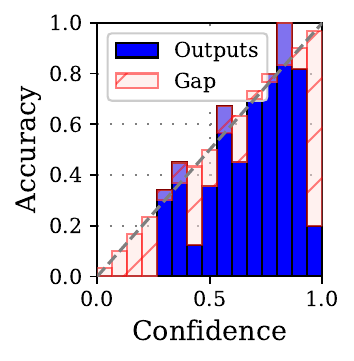} }}
	\subfloat[\scriptsize{Uncalibrated}]{{\includegraphics[width=0.2\textwidth]{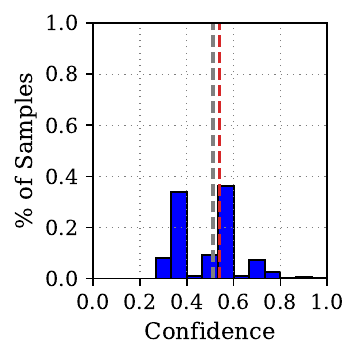} }}
	\subfloat[\scriptsize{TS ($\tau=0.83$)}]{{\includegraphics[width=0.2\textwidth]{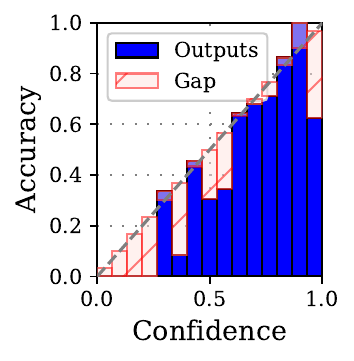} }}
    \subfloat[\scriptsize{TS ($\tau=0.83$)}]{{\includegraphics[width=0.2\textwidth]{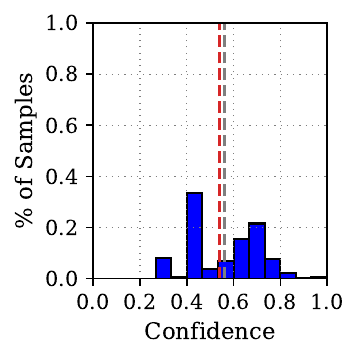} }} \\
	\subfloat[\scriptsize{Uncalibrated}]{{\includegraphics[width=0.2\textwidth]{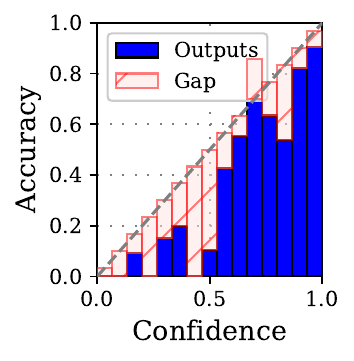} }}
	\subfloat[\scriptsize{Uncalibrated}]{{\includegraphics[width=0.2\textwidth]{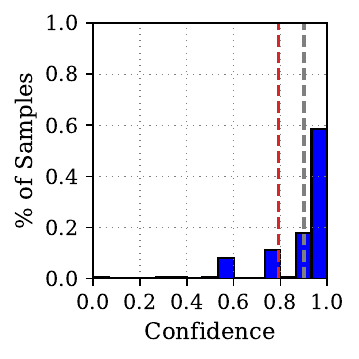} }}
	\subfloat[\scriptsize{TS ($\tau=1.67$)}]{{\includegraphics[width=0.2\textwidth]{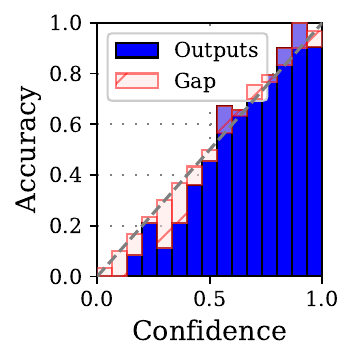} }}
	\subfloat[\scriptsize{TS ($\tau=1.67$)}]{{\includegraphics[width=0.2\textwidth]{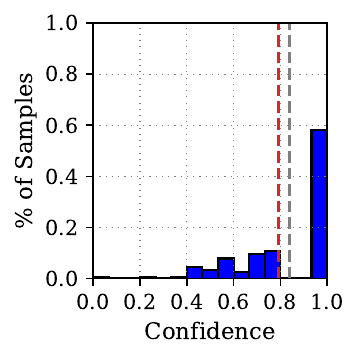} }} \\

	\caption{Claude v3 ($T=0.0$) reliability diagrams and confidence histograms for the test splits of the IMDB (top row), Emotion (middle row) and Amazon massive (bottom row) datasets. $\tau$ is tuned with TS on each corresponding validation dataset.}
\label{fig:calibration-claude-v3-T0}
\end{figure}

\begin{figure}[htbp]
	\centering
	\subfloat[\scriptsize{Uncalibrated}]{{\includegraphics[width=0.2\textwidth]{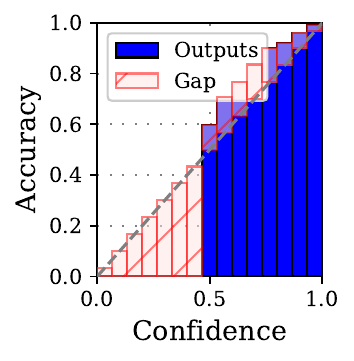} }}
	\subfloat[\scriptsize{Uncalibrated}]{{\includegraphics[width=0.2\textwidth]{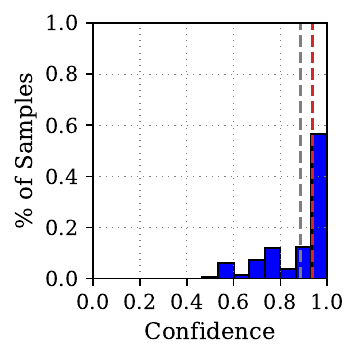} }}
	\subfloat[\scriptsize{TS ($\tau=0.63$)}]{{\includegraphics[width=0.2\textwidth]{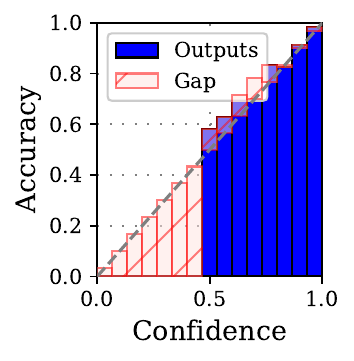} }}
	\subfloat[\scriptsize{TS ($\tau=0.63$)}]{{\includegraphics[width=0.2\textwidth]{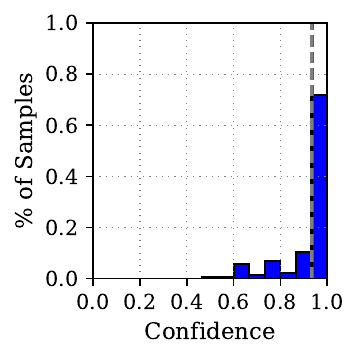} }} \\
	\subfloat[\scriptsize{Uncalibrated}]{{\includegraphics[width=0.2\textwidth]{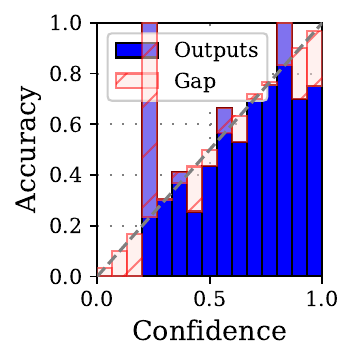} }}
	\subfloat[\scriptsize{Uncalibrated}]{{\includegraphics[width=0.2\textwidth]{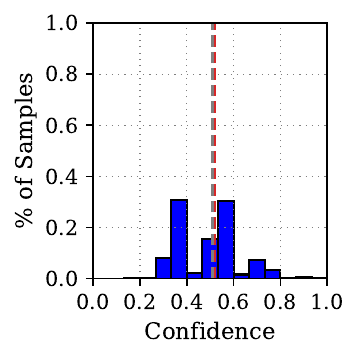} }}
	\subfloat[\scriptsize{TS ($\tau=0.88$)}]{{\includegraphics[width=0.2\textwidth]{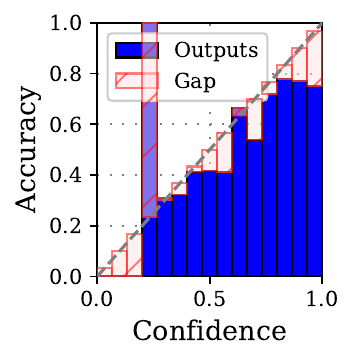} }}
    \subfloat[\scriptsize{TS ($\tau=0.88$)}]{{\includegraphics[width=0.2\textwidth]{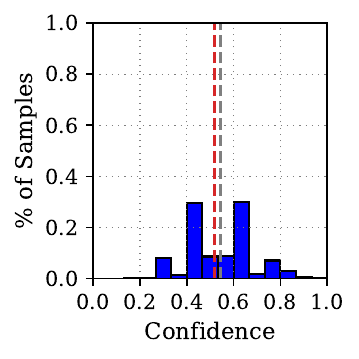} }} \\
	\subfloat[\scriptsize{Uncalibrated}]{{\includegraphics[width=0.2\textwidth]{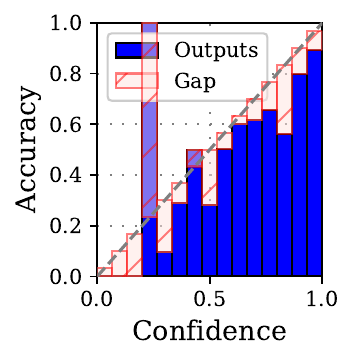} }}
	\subfloat[\scriptsize{Uncalibrated}]{{\includegraphics[width=0.2\textwidth]{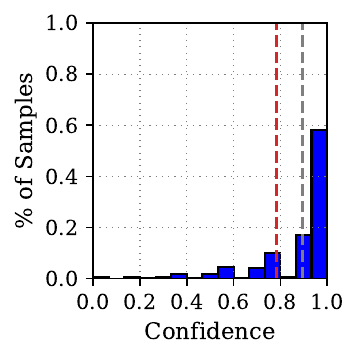} }}
	\subfloat[\scriptsize{TS ($\tau=1.33$)}]{{\includegraphics[width=0.2\textwidth]{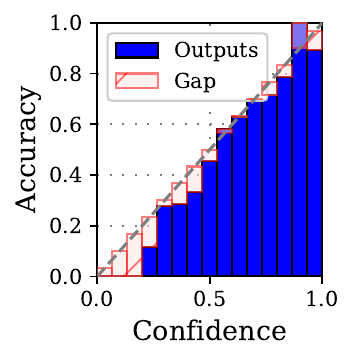} }}
	\subfloat[\scriptsize{TS ($\tau=1.33$)}]{{\includegraphics[width=0.2\textwidth]{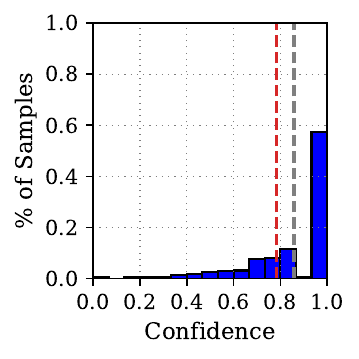} }} \\

	\caption{Claude v3 ($T=1.0$) reliability diagrams and confidence histograms
    for the test splits of the IMDB (top row), Emotion (middle row) and Amazon
massive (bottom row) datasets. $\tau$ is tuned with TS on each corresponding validation dataset.}
\label{fig:calibration-claude-v3-T1}
\end{figure}

\begin{figure}[htbp]
	\centering
	\subfloat[\scriptsize{Uncalibrated}]{{\includegraphics[width=0.2\textwidth]{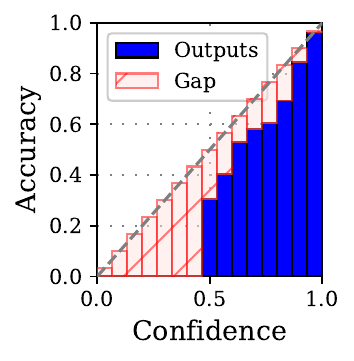} }}
	\subfloat[\scriptsize{Uncalibrated }]{{\includegraphics[width=0.2\textwidth]{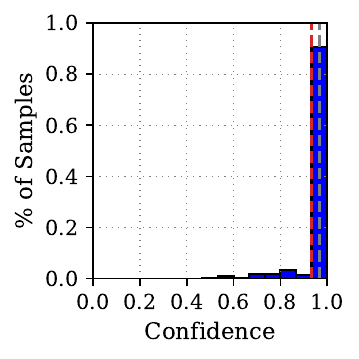} }}
	\subfloat[\scriptsize{TS ($\tau=1.39$)}]{{\includegraphics[width=0.2\textwidth]{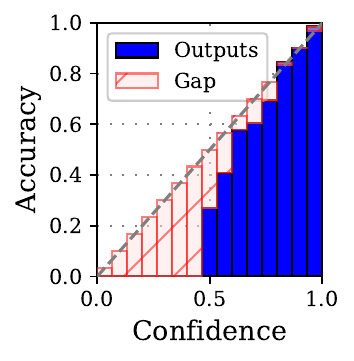} }}
	\subfloat[\scriptsize{TS ($\tau=1.39$)}]{{\includegraphics[width=0.2\textwidth]{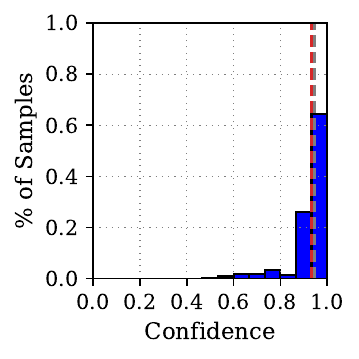} }} \\
	\subfloat[\scriptsize{Uncalibrated}]{{\includegraphics[width=0.2\textwidth]{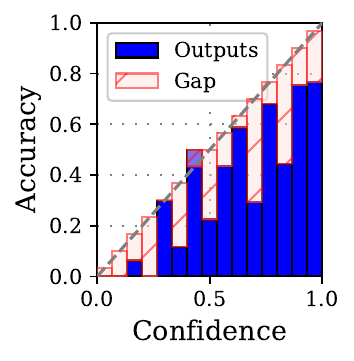} }}
	\subfloat[\scriptsize{Uncalibrated}]{{\includegraphics[width=0.2\textwidth]{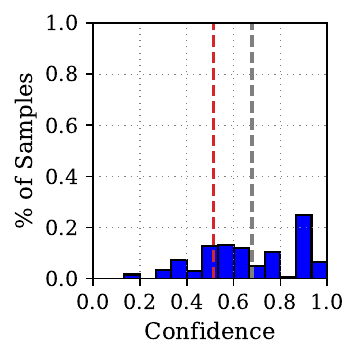} }}
	\subfloat[\scriptsize{TS ($\tau=1.75$)}]{{\includegraphics[width=0.2\textwidth]{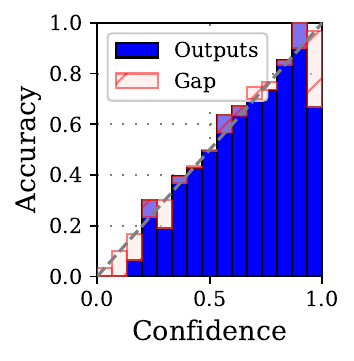} }}
    \subfloat[\scriptsize{TS ($\tau=1.75$)}]{{\includegraphics[width=0.2\textwidth]{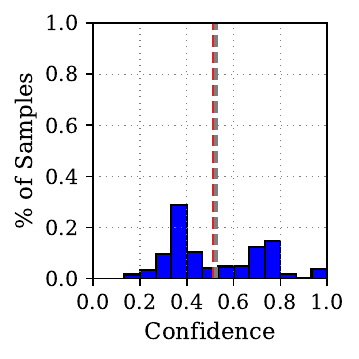} }} \\
	\subfloat[\scriptsize{Uncalibrated}]{{\includegraphics[width=0.2\textwidth]{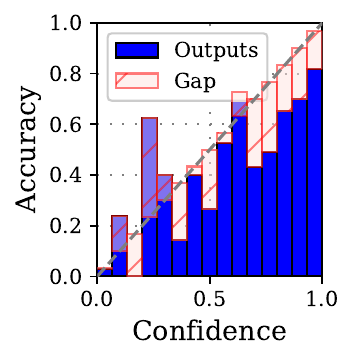} }}
	\subfloat[\scriptsize{Uncalibrated}]{{\includegraphics[width=0.2\textwidth]{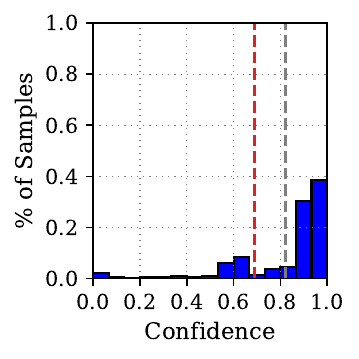} }}
	\subfloat[\scriptsize{TS ($\tau=1.65$)}]{{\includegraphics[width=0.2\textwidth]{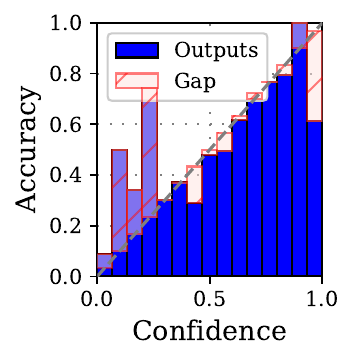} }}
	\subfloat[\scriptsize{TS ($\tau=1.65$)}]{{\includegraphics[width=0.2\textwidth]{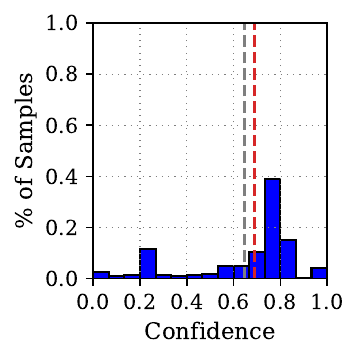} }} \\

	\caption{Mixtral ($T=0.0$) reliability diagrams and confidence histograms for the test splits of the IMDB (top row), Emotion (middle row) and Amazon massive (bottom row) datasets. $\tau$ is tuned with TS on each corresponding validation dataset.}
\label{fig:calibration-claude-v3-T0}
\end{figure}

\begin{figure}[htbp]
	\centering
	\subfloat[\scriptsize{Uncalibrated}]{{\includegraphics[width=0.2\textwidth]{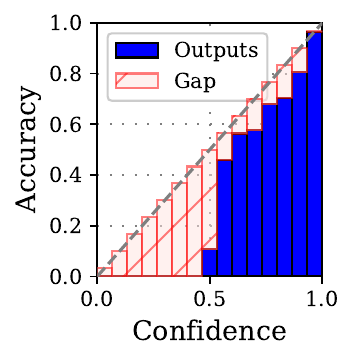} }}
	\subfloat[\scriptsize{Uncalibrated }]{{\includegraphics[width=0.2\textwidth]{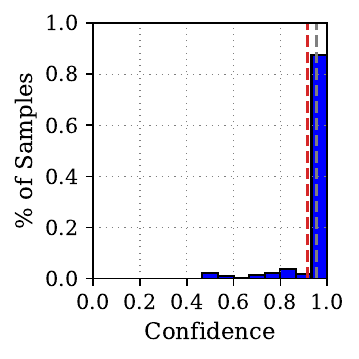} }}
	\subfloat[\scriptsize{TS ($\tau=1.42$)}]{{\includegraphics[width=0.2\textwidth]{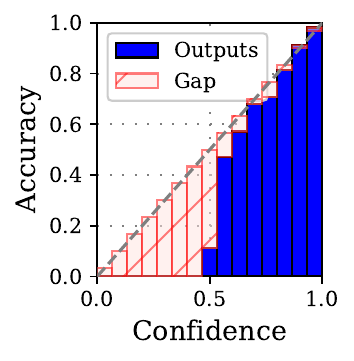} }}
	\subfloat[\scriptsize{TS ($\tau=1.42$)}]{{\includegraphics[width=0.2\textwidth]{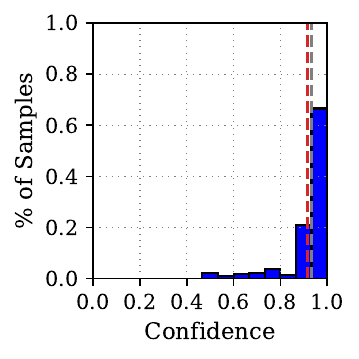} }} \\
	\subfloat[\scriptsize{Uncalibrated}]{{\includegraphics[width=0.2\textwidth]{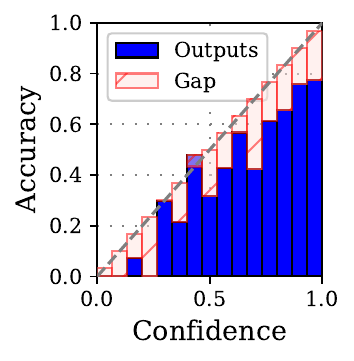} }}
	\subfloat[\scriptsize{Uncalibrated}]{{\includegraphics[width=0.2\textwidth]{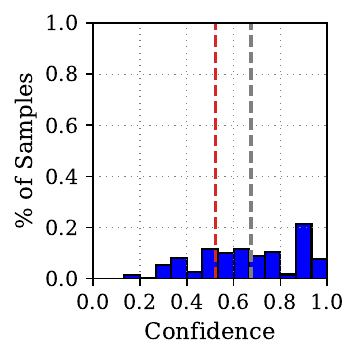} }}
	\subfloat[\scriptsize{TS ($\tau=1.73$)}]{{\includegraphics[width=0.2\textwidth]{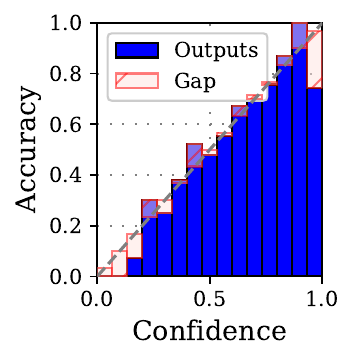} }}
    \subfloat[\scriptsize{TS ($\tau=1.73$)}]{{\includegraphics[width=0.2\textwidth]{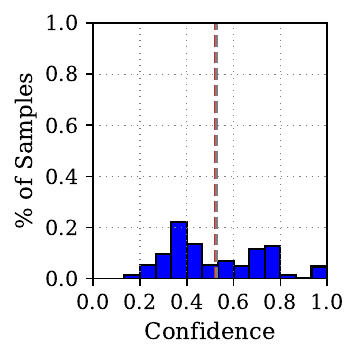} }} \\
	\subfloat[\scriptsize{Uncalibrated}]{{\includegraphics[width=0.2\textwidth]{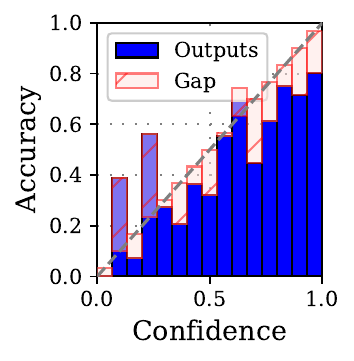} }}
	\subfloat[\scriptsize{Uncalibrated}]{{\includegraphics[width=0.2\textwidth]{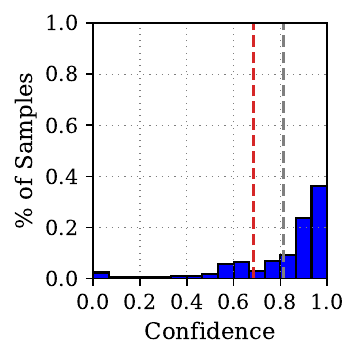} }}
	\subfloat[\scriptsize{TS ($\tau=1.64$)}]{{\includegraphics[width=0.2\textwidth]{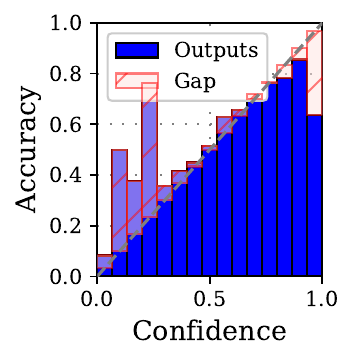} }}
	\subfloat[\scriptsize{TS ($\tau=1.64$)}]{{\includegraphics[width=0.2\textwidth]{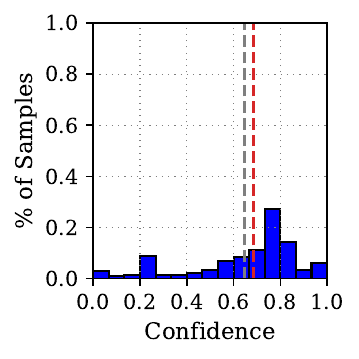} }} \\

	\caption{Mixtral ($T=1.0$) reliability diagrams and confidence histograms for the test splits of the IMDB (top row), Emotion (middle row) and Amazon massive (bottom row) datasets. $\tau$ is tuned with TS on each corresponding validation dataset.}
\label{fig:calibration-claude-v3-T0}
\end{figure}

\end{document}